\documentclass{article}
\usepackage[numbers, compress]{natbib}
\usepackage[preprint]{neurips_2023}
\usepackage[utf8]{inputenc} 
\usepackage[T1]{fontenc}    
\usepackage{hyperref}       
\usepackage{url}            
\usepackage{booktabs}       
\usepackage{amsfonts}       
\usepackage{nicefrac}       
\usepackage{microtype}      
\usepackage[dvipsnames]{xcolor}         

\usepackage{amsmath,amsfonts,bm}

\def\eqref#1{equation~\ref{#1}}

\def\1{\bm{1}}

\DeclareMathAlphabet{\mathsfit}{\encodingdefault}{\sfdefault}{m}{sl}
\SetMathAlphabet{\mathsfit}{bold}{\encodingdefault}{\sfdefault}{bx}{n}

\usepackage{url}
\usepackage{microtype}
\usepackage{graphicx}
\usepackage{booktabs} 
\usepackage{floatrow}
\usepackage{hyperref}
\usepackage{tikz-cd}
\usepackage{amssymb, amsmath, amsthm, amsfonts}
\usepackage{soul}
\usepackage{tikz}
\usepackage{subfigure}
\usepackage{wrapfig}
\usepackage[ruled,vlined]{algorithm2e}
\usepackage{algorithmic}
\usepackage[utf8]{inputenc} 
\usepackage[T1]{fontenc}    
\usepackage{hyperref}       
\usepackage{url}            
\usepackage{nicefrac}       
\usepackage{bbm}
\usepackage{xcolor}
\usepackage{color}
\usepackage{threeparttable}
\usepackage{caption}

\usepackage{multibib}

\definecolor{c1}{RGB}{255,75,0}
\definecolor{c2}{RGB}{0,180,255}
\definecolor{LightCyan}{rgb}{0.88,1,1}
\definecolor{Gray}{gray}{0.9}
\usepackage{diagbox}
\usepackage[
    separate-uncertainty = true,
    multi-part-units = repeat
]{siunitx}
\usepackage{microtype}
\usepackage{booktabs} 
\usepackage{multirow}
\usepackage{pifont}
\usepackage{colortbl}
\usepackage{float}
\usepackage{paralist,tabularx}
\usepackage{array}
\newcolumntype{?}{!{\vrule width 1pt}}
\usetikzlibrary{fit,positioning}

\colorlet{mylinkcolor}{violet}
\colorlet{mycitecolor}{RoyalBlue}
\colorlet{myurlcolor}{RoyalPurple}
\hypersetup{
  citecolor  = mycitecolor,
  linkcolor = myurlcolor,
  urlcolor = myurlcolor,
  colorlinks = true
}

\definecolor{myredcolor}{RGB}{215,48,39}
\definecolor{mygreencolor}{RGB}{26,152,80}
\definecolor{goldcolor}{RGB}{255,165,0}

\makeatletter
\setbox0\hbox{$\xdef\scriptratio{\strip@pt\dimexpr
\numexpr(\sf@size*65536)/\f@size sp}$}
\newcommand{\scriptveryshortarrow}[1][3pt]{{%
    \vcenter{\hbox{\rule[\scriptratio\dimexpr-.2pt\relax]
    {\scriptratio\dimexpr#1\relax}{\scriptratio\dimexpr.4pt\relax}}}%
    \mkern-4mu\hbox{\let\f@size\sf@size\usefont{U}{lasy}{m}{n}\symbol{41}}}}
\usepackage{pifont}

\newcommand{\cmark}{\textcolor{black}{\ding{51}}}
\newcommand{\xmark}{\textcolor{black}{\ding{55}}}

\newtheorem{definition}{Definition}

\usepackage[utf8]{inputenc} 
\usepackage[T1]{fontenc}    
\usepackage{courier}
\usepackage{hyperref}       
\usepackage{url}            

\usepackage{booktabs}       
\usepackage{amsfonts}       
\usepackage{amsmath} 
\usepackage{nicefrac}       
\usepackage{wrapfig}
\usepackage{multirow}

\usepackage{tikz}
\usetikzlibrary{fit,matrix,positioning, 
decorations.pathreplacing,calc,decorations,arrows,shadows,patterns}
\usetikzlibrary{arrows,bayesnet,fit,positioning,shapes,matrix}
\newdimen\arrowsize
\pgfarrowsdeclare{arcsq}{arcsq}
{
	\arrowsize=0.2pt
	\advance\arrowsize by .5\pgflinewidth
	\pgfarrowsleftextend{-4\arrowsize-.5\pgflinewidth}
	\pgfarrowsrightextend{.5\pgflinewidth}
}
{
	\arrowsize=0.8pt
	\advance\arrowsize by .5\pgflinewidth
	\pgfsetdash{}{0pt} 
	\pgfsetroundjoin   
	\pgfsetroundcap    
	\pgfpathmoveto{\pgfpoint{0\arrowsize}{0\arrowsize}}
	\pgfpatharc{-90}{-140}{4\arrowsize}
	\pgfusepathqstroke
	\pgfpathmoveto{\pgfpointorigin}
	\pgfpatharc{90}{140}{4\arrowsize}
	\pgfusepathqstroke
}

\usepackage{amsmath}
\usepackage{amssymb}
\usepackage{mathtools}
\usepackage{amsthm}
\usepackage{textcomp}

\usepackage{xcolor,colortbl}
\definecolor{ourmethod}{gray}{0.93}
\definecolor{myredcolor}{RGB}{215,48,39}
\definecolor{mygreencolor}{RGB}{26,152,80}
\usepackage{pifont}

\definecolor{dred}{rgb}{0.8, 0.0, 0.0}

\usepackage{enumitem}

\usepackage[utf8]{inputenc} 
\usepackage[T1]{fontenc}    
\usepackage{hyperref}       
\usepackage{url}            
\usepackage{booktabs}       
\usepackage{amsfonts}       
\usepackage{nicefrac}       
\usepackage{microtype}      
\usepackage{xcolor}         

\newcommand{\ours}{\textbf{\texttt{DAFT}} }

\title{Learning Dynamic Attribute-factored World Models for Efficient Multi-object Reinforcement Learning}

\author{
    Fan Feng$^1$, Sara Magliacane$^{2,3}$\\
    $^1$City University of Hong Kong
    $^2$University of Amsterdam
    $^3$MIT-IBM Watson AI Lab\\
    {\small \texttt{\{ffeng1017,sara.magliacane\}@gmail.com}}}

\begin{document}

\maketitle

\begin{abstract}
In many reinforcement learning tasks, the agent has to learn to interact with many objects of different types and generalize to unseen combinations and numbers of objects. Often a task is a composition of previously learned tasks (e.g. block stacking).
These are examples of \emph{compositional generalization}, in which we compose object-centric representations to solve complex tasks. Recent works have shown the benefits of object-factored representations and hierarchical abstractions for improving sample efficiency in these settings. On the other hand, these methods do not fully exploit the benefits of factorization in terms of object attributes.
In this paper, we address this opportunity and introduce the Dynamic Attribute FacTored RL (DAFT-RL) framework. In DAFT-RL, we leverage object-centric representation learning to extract objects from visual inputs. We learn to classify them in classes and infer their latent parameters. For each class of object, we learn a class template graph that describes how the dynamics and reward of an object of this class factorize according to its attributes. We also learn an interaction pattern graph that describes how objects of different classes interact with each other at the attribute level. Through these graphs and a dynamic interaction graph that models the interactions between objects, we can learn a policy that can then be directly applied in a new environment by just estimating the interactions and latent parameters.
We evaluate DAFT-RL in three benchmark datasets and show our framework outperforms the state-of-the-art in generalizing across unseen objects with varying attributes and latent parameters, as well as in the composition of previously learned tasks.
\end{abstract}


\section{Introduction}


Model-based reinforcement learning (MBRL) and world models \cite{hafner2019learning, hafnerdream, micheli2022transformers} have demonstrated improved performance in many RL tasks by providing better sample efficiency. 
However, most world models focus only on modeling a single object or holistic modeling over the environment, while in real-world tasks, we often have environments with multiple objects that interact, and we are interested to generalize to unseen combinations and numbers of objects. 
In recent years, there have been several studies exploring and learning object-oriented environment models or policy models~\cite{li2020towards, zadaianchuk2021selfsupervised, zadaianchuk2022self, mambelli2022compositional, Kossen2020Structured, watters2019cobra, goyal2021recurrent, choi2023unsupervised, chang2023hierarchical, yoon2023investigation, zhao2022toward, zhou2022policy, biza2022factored, nakanointeraction} and tackling the problem of \emph{compositional} or combinatorial generalization, in which we consider combining the modeling of multiple objects or tasks to solve a new task. 

Although these methods have effectively leveraged object-centric and object-factored representations in RL, and thus improved the sample efficiency in multi-object settings, they have not fully exploited the benefits of factorization in terms of object attributes. Often an object's transition and reward functions are influenced only by a sparse subset of attributes, e.g. an object's position and reward are affected by its previous position, but not by its appearance or activation state. In these environments, interactions between objects are often sparse, both in time and in terms of which attributes are affected, e.g. the position of a box is affected by the position of another box at the timestep in which they collide, but not directly by the other object's friction coefficient. Additionally, objects of the same type share similar factored dynamics, modulated by an object-specific latent parameter, while objects of different types might instead have different attributes, dynamics, and rewards.

In this paper, we propose Dynamic Attribute FacTored
RL (DAFT-RL), a framework that learns a fine-grained attribute-factored representation across objects, including a dynamic graph for modeling interactions between objects. As part of this framework, we propose a model, DAFT-MDP, that builds on Factored (PO)MDPs \cite{, sallans2004reinforcement,kearns1999efficient,boutilier2000stochastic,guestrin2003efficient,hallak2015off,factoredPOMDP}, Relational MDPs \cite{gardiol2003envelope,guestrin2003generalizing,van2005survey} and especially Object-Oriented (PO)MDPs \cite{diuk2008object,pmlr-v32-scholz14, OO-POMDP}, but focuses on a more fine-grained factorization at the attribute level and dynamic graphs. We implement our framework as a structured and sequential generative model by combining it with state-of-the-art object-centric representation learning \cite{locatello2020object,eslami2016attend} for extracting objects and attributes from visual inputs, factored adaptation approaches inspired by the causality literature \cite{huang2021adarl,feng2022factored} for estimating the factored dynamics and reward, soft attention networks \cite{vaswani2017attention} for action binding \cite{biza2022binding} and (dynamic) Neural Relational Inference \cite{kipf2018neural, graber2020dynamic} for modeling interactions. Our framework allows us to learn a policy on a set of source environments that can successfully generalize to new environments with unseen combinations of objects with different latent parameters  (possibly with unseen values) and types, as well as to combinations of previously learned tasks on different objects, without any further policy learning. We show the benefits of DAFT-RL in three benchmark datasets for compositional generalization, where it outperforms the baselines.

\section{Dynamic Attribute-FacTored MDPs (DAFT-MDP)}
\label{sec2}
We formalize our assumptions by introducing our DAFT-MDP model, which is an extension with class template graphs, interaction patterns, and interaction graphs of factored (PO)MDPs \cite{sallans2004reinforcement,kearns1999efficient,boutilier2000stochastic,guestrin2003efficient,hallak2015off, factoredPOMDP}. This extension takes inspiration from Relational MDPs \cite{gardiol2003envelope,guestrin2003generalizing,van2005survey} and their literature, especially Object-Oriented (PO)MDPs \cite{diuk2008object,pmlr-v32-scholz14, OO-POMDP}, but we propose a more fine-grained factorization of the transition and reward at the object attribute level, based on estimating template and dynamic interaction graphs.

Intuitively, we will consider learning a policy that can generalize across different environments that vary in the number and characteristics of their objects. We will assume that each environment is composed of multiple \emph{objects}, each of a specific type, or \emph{class}. Each object has several observable \emph{attributes} (e.g. position, velocity) and some latent constant \emph{parameters} (e.g. an object-specific friction coefficient). Objects of the same class will have the same attributes, the same transition and reward functions, but can differ in the values of the attributes (e.g. they are at different positions) and in the value of the latent parameters (e.g. they have different friction coefficients). We will assume that the transition and reward functions can be \emph{factored} in terms of attributes and that for a given attribute only a sparse subset of other attributes influences these functions.
The objects can interact with each other, which might influence their dynamics. We will assume that these \emph{interactions} are \emph{sparse}, both in time and in terms of the effect on the attributes of each object, and that all objects in a class have the same \emph{interaction pattern} in terms of how the attributes interact with objects of another class. In each environment, we will assume that an action has only an effect on one object at a time. 

\begin{figure}[t]
    \centering
\includegraphics[width=0.98\linewidth]{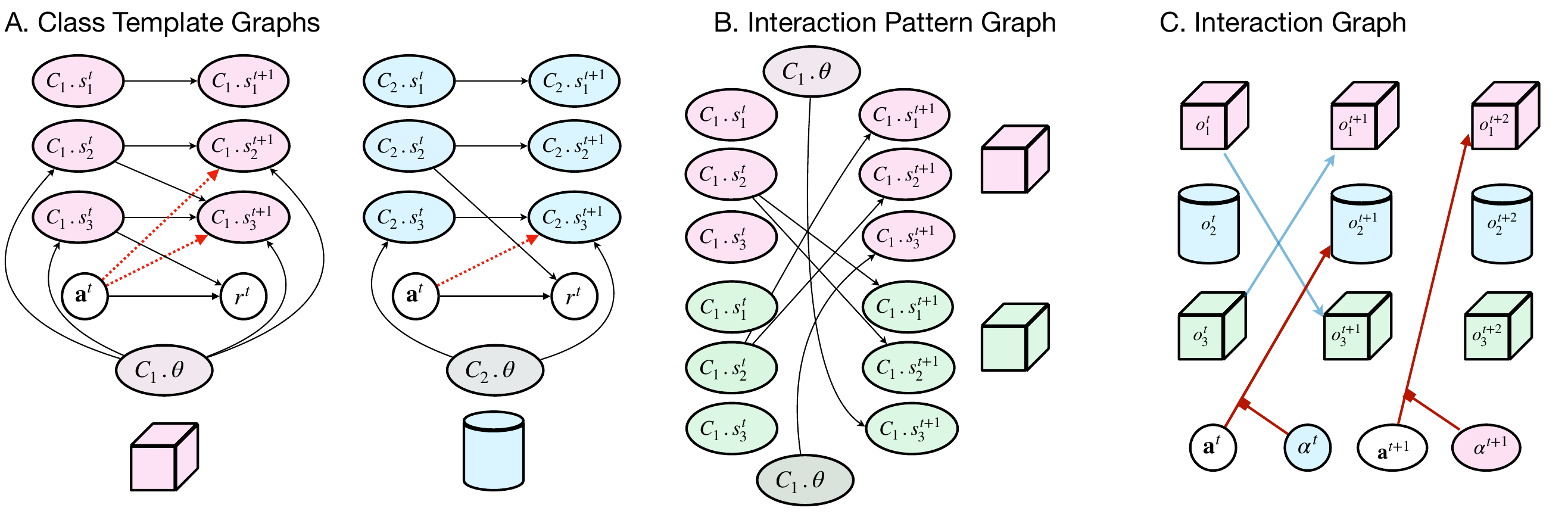}
    \caption{The graphical representation of DAFT-MDP. The colors denote the attributes for an object or a class, the red dashed lines denote edges that can be switched on or off at different timesteps. }
    \label{fig:factored_graph}
\end{figure}

We formalize these assumptions in the following.
We start by defining our class system, then we describe three types of graphs (class template graphs, interaction patterns, and dynamic interaction graphs in Fig.~\ref{fig:factored_graph}) that describe how dynamics and reward factorize based on the classes, attributes, and interactions, and finally provide a formal definition of a Dynamic Attribute-FacTored MDP.

\paragraph{Class system, attributes, and objects.} We assume a known set of classes $\mathcal{C} = \{C_1, \dots, C_k \}$ of objects, where each class $C_j$ describes a set of observable attributes $\{C_j.s_1, \dots, C_j.s_n \}$, which we assume for simplicity are the same number in each class. We assume that each class has a set of latent constant parameters $C_j.\boldsymbol{\theta}$, that represent physical properties of the object that can vary across different objects of the same type.
For example, in one of the benchmarks \cite{zhou2022policy}, we consider two types of objects  $\mathcal{C} = \{\textsc{box}, \textsc{sw}\}$, boxes and switches, with class attributes $\{\textsc{box}.\mathrm{pos}, \textsc{box}.\mathrm{vel}, \textsc{box}.\boldsymbol{\theta} 
 \}$ representing the position, velocity, and friction coefficient of a box, and $\{\textsc{sw}.\mathrm{pos}, \textsc{sw}.\mathrm{active}
 \}$, representing the position and activation state of a switch. 
The class system specifies a template for a set of environments that can vary in the number and characteristics of the objects. Each environment has a fixed set of objects $\mathcal{O} = \{o_1, \dots, o_m\}$, where each object is an instance of a class in $\mathcal{C}$, which we denote as $o_i \in C_j$ for $i \in \{1,\dots, m \}, j \in \{1, \dots, k\}$. We denote the class of an object $o_i$ with $C(i)$.
For example, using the class system from the previous example, we can represent a source environment as  $\mathcal{O}' = \{o_1, o_2\}$ where $o_1 \in \textsc{box}$ and $o_2 \in \textsc{sw}$, and a target environment as $\mathcal{O}{''} = \{o_1, o_2, o_3\}$, where we add $o_3 \in \textsc{box}$ to the original objects. 
 For each object $o_i \in C_j$ and timestep $t=\{1, \dots, T\}$, we denote its attributes at time $t$ as $\mathbf{o}_i^t=\{o_i.s_1^t, \dots, o_i.s_n^t\}$, which are instantiations of the class attributes $\{C_j.s_1, \dots, C_j.s_n \}$, and its constant parameters as $o_i.\boldsymbol{\theta}$. In our example, for box $o_1$ the attributes  $o_1.s_1^t$, $o_1.s_2^t$ and $o_1.\boldsymbol{\theta}$ are its position and velocity at time $t$ and its friction coefficient, while for the switch $o_2$ the attributes $o_2.s_1^t$  and $o_2.s_2^t$ are its position and activation at time $t$.

\paragraph{States, actions, transitions, and rewards.} We define the state at time $t$ as the collection of all the object states, i.e., $\mathbf{s^t}=\{\mathbf{o_1^t}, \dots, \mathbf{o_m^t}\}$ with domain $\mathcal{S}$. We collect all object-specific latent parameters in a global parameter $\boldsymbol{\theta} = \{ o_1.\boldsymbol{\theta}, \dots,  o_m.\boldsymbol{\theta} \}$ with domain $\Theta$.
We define the space of actions $\mathcal{A}$ and use $\mathbf{a}^t \in \mathcal{A}$ to denote the action at time $t$. We denote the transition probability as $p(\mathbf{s^{t+1}} | \mathbf{s^{t}}, \mathbf{a}^t, \boldsymbol{\theta})$. The reward at time $t$ is denoted as $r^t$ and the reward probability is denoted as $p(r^t | \mathbf{s^{t}}, \mathbf{a}^t,\boldsymbol{\theta})$. The transition and reward probabilities are factorized according to the \emph{class template graphs}, \emph{interaction pattern graphs} and \emph{interaction graphs} that we can learn from data, and that we will introduce below.

\paragraph{Class template graphs (Fig.~\ref{fig:factored_graph}A).} We assume that all objects within the same class share the same factorization in terms of how their own attributes, latent parameters and actions influence their dynamics and rewards. For example, all boxes will have a similar relationship between their position, velocity, and latent friction parameters, since they follow the same physical laws. Similarly, all switches will share similar dynamics, which will be different from the boxes. We describe these relationships as Dynamic Bayesian Networks (DBNs) \cite{Murphy_DBN}, which are graphical models that describe the template for the relations between two contiguous timesteps, $t$ and $t+1$, and assume this structure is time-invariant.
In particular, for each class $C_j$, we learn a DBN $\mathcal{G}_{C_j}$ over the nodes $\{  C_j.s_1^t, \dots,C_j.s_n^t, C_j.s_1^{t+1}, \dots, C_j.s_n^{t+1}, C_j.\boldsymbol{\theta}, \mathbf{a}^t, r^t \}$ that represents a template for the instance graph between the attributes for each object of this class $\mathcal{G}^t_{o_i}$ for $o_i \in C_j$. In particular, the edges between the action $a^t$ and the attributes at timestep $t+1$ can be switched on or switched off at different timesteps, since as we will see in the description of the interaction graphs, they represent the interaction pattern of the agent with any specific object. We show an example of two DBNs for two classes of objects (boxes and switches) in Fig.~\ref{fig:factored_graph}a. In this example, $C_1.s_2^t$ influences $C_1.s_2^{t+1}$ and $C_1.s_3^{t+1}$, but it's not influenced by $C_1.s_3^{t}$. The reward $r^t$ is only influenced by $C_1.s_3^{t}$ and the action will only have an effect on $C_1.s_2^{t+1}$ and $C_1.s_3^{t+1}$. Moreover, $C_2.s_2^t$ influences only $C_2.s_2^{t+1}$ and $r^t$.

\paragraph{Interaction pattern graphs (Fig.~\ref{fig:factored_graph}B).} When two objects interact, it often happens that only some of the attributes of each object after the other. For example, when two boxes collide, their positions and velocities might change, but their masses will not. Therefore, we will assume that from an attribute perspective, the interplay between objects during interaction is also factored and sparse. Additionally, we assume that the interactions between two objects follow patterns based on the classes of the objects. For example, the attribute of a box will always interact in the same way with the attributes of a switch, regardless of the specific object. In particular, for any pair of classes $C_i$ and $C_j$ (possibly also with $i=j$), we learn a DBN $\mathcal{G}_{C_i, C_j}$ over the nodes $\{  \{C_i.s_l^t \}_{l=1}^n,\{C_i.s_l^{t+1} \}_{l=1}^n, C_i.\boldsymbol{\theta}, \{C_j.s_l^t \}_{l=1}^n,\{C_j.s_l^{t+1} \}_{l=1}^n, C_j.\boldsymbol{\theta} \}$. We show an example of a DBN describing how boxes interact with other boxes in Fig.~\ref{fig:factored_graph}b. In this case, the interaction is limited to $C_1.s_2^t$ from one object influencing the $C_1.s_1^{t+1}$ and $C_1.s_2^{t+1}$ from the other object, and the latent parameters $C_1.\boldsymbol{\theta}$ from each object influencing $C_1.s_3^{t+1}$ from the other object. While the interaction patterns are time-invariant, we assume that for each pair of objects, the interactions are only switched on at some points in time, as we will now describe through the interaction graphs.

\paragraph{Dynamic interaction graph (Fig.~\ref{fig:factored_graph}C).} The class template and interaction pattern graphs that we just described can model the general behavior of the classes of objects in a static, time-invariant way. On the other hand, in many multi-object environments object interactions occur sparsely, the pairs of interacted objects are not fixed, and the action has an effect only on a limited number of objects at any given time (e.g. we will assume only one for simplicity). We therefore propose to model these interactions between objects as a dynamic graph  $\mathcal{G}_{\mathrm{inter}} = \{\mathcal{G}^t_{\mathrm{inter}}\}_{t=1}^T$ at the object level, which is a sequence of graphs $\mathcal{G}^t_{\mathrm{inter}}$ with edges from a subset of $\{o_1^t, \dots, o_m^t, \mathbf{a}^t\}$ to a subset of $\{o_1^{t+1}, \dots, o_m^{t+1}\}$. Each edge $o_i^t \to o_j^{t+1}$ represents an interaction between an object $o_i$ and an object $o_j$. This interaction is instantiated 
in an instance interaction graph $\mathcal{G}^t_{o_i, o_j}$, 
following the interaction pattern graph $\mathcal{G}_{C_i, C_j}$ for each pair of objects $o_i \in C_i$ and $o_j \in C_j$.
We also learn an \emph{object selector} $\alpha^t \in \{1, \dots, m\}$ that selects the single object on which the action $\mathbf{a}^t$ has an effect at timestep $t$, which we represent with an edge from $\mathbf{a}^t$ to the selected object $o^t_{\alpha^t}$ in $\mathcal{G}^t_{\mathrm{inter}}$ .  In particular, this selection (or \emph{action binding}) activates the edges from $\mathbf{a}^t$ to the object attributes described in the class template graph $\mathcal{G}_{C_j}$ in the instantiated version for object $o_i \in C_j$, which we denote $\mathcal{G}^t_{o_i}$.
The graph $\mathcal{G}_{\mathrm{inter}}$ is dynamic because the edges of each graph in the sequence can change at each timestep $t = 1, \dots, T$. 
We show an example of interactions between three objects at different timesteps in Fig.~\ref{fig:factored_graph}c. In this example, at timestep $t$ there is an interaction between $o_1$ and $o_3$, which will follow the interaction pattern presented in Fig.~\ref{fig:factored_graph}b, i.e. $o_1.s_2^t \to \{o_3.s_1^{t+1}, o_3.s_2^{t+1}\}$ and $o_1.\boldsymbol{\theta}^t \to o_3.s_3^{t+1}$, and viceversa for $o_3$ towards $o_1$.
Moreover, $\mathbf{a}^t$ has an effect on $o_2$, and specifically $o_2.s_3^{t+1}$, following the class template graph presented in Fig.~\ref{fig:factored_graph}a. Instead, at timestep $t+1$, there are no interactions between the objects and $\mathbf{a}^{t+1}$ has an effect on $o_1$, specifically $o_1.s_2^{t+2}$ and $o_1.s_3^{t+2}$. For completeness, in App.~\ref{app:fullexample} we provide an example of this graph combined with the instantiated class template graphs for each object $\mathcal{G}^{o_1}, \mathcal{G}^{o_2}$ and $\mathcal{G}^{o_3}$ for three timesteps $t, t+1, t+2$, as well as the instantiated interaction pattern graph $\mathcal{G}^{o_1, o_3}$ that is switched on at timestep $t$. 

\paragraph{Our modeling assumptions. }
Now that we have introduced all of the graphical structures that we will need, we can describe our assumptions as a Dynamic Attribute-FacTored Markov Decision Process (DAFT-MDP). We will assume that the class system is fixed and that the objects can vary in each environment, as can their interaction graphs. Under this assumption, a DAFT-MDP defines a family of MDPs, that is parametrized in the objects $\mathcal{O}$ and their dynamic interaction graph $\mathcal{G}_{\mathrm{inter}}$.

\begin{definition}[DAFT-MDP] \label{def:daft-mdp}
 A \emph{Dynamic Attribute-FacTored Markov Decision Process (DAFT-MDP)} is a tuple $\left(\mathcal{C}, \mathcal{O},  \Theta, \mathcal{A},  \mathcal{G}, \mathbb{P}_s, \mathcal{R},  \gamma \right)$, where $\mathcal{C}$ is the set of classes, $\mathcal{O}$ is the set of objects, $\Theta$ is the space of the constant latent parameters, $\mathcal{A}$ the action space, $\mathbb{P}_s$ is the transition distribution, $\mathcal{R}$ is the reward function and $\mathcal{G}$ is a set of graphs that contains the collection of class template graphs for each class  $\{\mathcal{G}_{C_j}\}_{C_j \in \mathcal{C}}$, the collection of interaction pattern graphs for each pair of classes $\{\mathcal{G}_{C_i, C_j}\}_{C_i, C_j \in \mathcal{C}}$ and the dynamic interaction graph $\mathcal{G}_{\mathrm{inter}}$, as defined previously.
 These graphs define the factorization of the transition distribution per object and per attribute, as follows:
 \begin{align*}
            \mathbb{P}_s (\boldsymbol{s}^{t+1} | \boldsymbol{s}^{t},\boldsymbol{\theta}, \boldsymbol{a}^{t} ) = \prod_{i=1}^m\prod_{l=1}^n \mathbb{P}_s \left( o_{i}. s_l^{t+1}| \mathrm{pa}_{\mathcal{G}^t_{o_i}} ( o_{i}. s_l^{t+1}), \{\mathrm{pa}_{\mathcal{G}^t_{o_i, o_k}} (o_{i}. s_l^{t+1}) \}_{o_k \to o_i \in \mathcal{G}^t_{\mathrm{inter}}}\right)
        \end{align*}
where $\boldsymbol{s}^{t}$ is the collection of all attributes of all objects at time $t$, $\boldsymbol{\theta} \in \Theta$ is the collection of all latent constant parameters for all objects, $\boldsymbol{a}^{t} \in \mathcal{A}$ is the action. In the right-hand term, $o_{i}.s_l^{t+1}$ is attribute $s_l$ of object $o_i$ at time $t+1$, while  $\mathrm{pa}_{\mathcal{G}^t_{o_i}} (o_{i}.s_l^{t+1})$ are the parents of the attribute $l$ for object $o_i$ based on the class template graph $\mathcal{G}_{C(i)}$, where $C(i)$ is the class of $o_i$, and where the action binding $\alpha^t$ activates any potential connections from $\mathbf{a}^t$. In the second term of the conditioning, we iterate over the objects $o_k$ that are interacting with $o_i$ at time $t$ in the dynamic interaction graph $o_k \to o_i \in \mathcal{G}^t_{\mathrm{inter}}$. For each of these objects $o_k$ we collect the attributes that interact with $o_i.s_l$ in the instance interaction pattern $\mathcal{G}^t_{o_i, o_k}$ based on interaction pattern graph $\mathcal{G}_{C(i), C(k)}$ for the respective classes $C(i)$ and $C(k)$. Similarly, we define the factorization of the reward function per object and per attribute as $\mathcal{R}( \mathbf{s}^t, \mathbf{a}^t, \boldsymbol{\theta}) = \mathcal{R}(\{\mathrm{pa}_{\mathcal{G}_{o_i}} (r^t)\}_{o_i \in \mathcal{O}})$, where for each object $o_i$ we collect all the attributes that have an edge to the reward in the instantiation of the class template graph.
\end{definition}
In the following, we assume that the classes $\mathcal{C}$ are known and fixed across environments, while the objects $\mathcal{O}$ can vary, as can the latent parameters $\boldsymbol{\theta}$. In the training phase, we will learn how to classify objects, the transition and reward functions based on the class template graphs $\{\mathcal{G}_{C_j}\}_{C_j \in \mathcal{C}}$ and the interaction patterns $\{\mathcal{G}_{C_i, C_j}\}_{C_i, C_j \in \mathcal{C}}$. In the testing phase, we will infer the class and latent parameters of each object, as well as the interactions between the objects in the dynamic interaction graph $\mathcal{G}^t_{\mathrm{inter}}$, which specify the transition and reward functions in the new environment.

\section{The DAFT-RL framework}
In the previous section, we introduced our model, DAFT-MDP. In this section we provide a framework for estimating DAFT-MDPs, leveraging them for policy learning in a set of source environments, and adapting the policy to a new target environment with different objects, without any additional policy learning. Our framework is divided in four steps: (i) offline class learning in single-object environments, (ii) offline interaction learning and latent parameter inference in multi-object environments, (iii)  policy learning and imagination in multi-object environments, and finally (iv) adaptation to a new multi-object environment. We present each step and its implementation in the following.

In all steps and all environments, if the input is an image, we extract the objects and their attributes $\mathbf{s}^t =\{\mathbf{o}_1^t,\mathbf{o}_2^t, \ldots,  \mathbf{o}_m^t\}$ from sequences of images $\mathbf{x}^t$  with pre-trained object-centric methods, e.g. SA~\cite{locatello2020object} and AIR~\cite{eslami2016attend}.
For symbolic inputs, we directly access the attributes of all objects $\mathbf{s}^t =\{\mathbf{o}_1^t,\mathbf{o}_2^t, \ldots,  \mathbf{o}_m^t\}$. For each set of objects $\mathcal{O}$, we learn to classify the objects into their classes $\mathcal{C}$ with supervised learning, which we describe in detail in App.~\ref{app:algorithm}.

\subsection{Step 1: Class learning in single-object environments.} 
\label{sec:step1}

In this step, we consider data from $m$ single-object environments with different objects and no agent interaction. 
In particular, for each class $C_j$ we collect the transitions for several objects $o_i \in C_j$ as $\{\mathbf{o}_i^t, \mathbf{a}_i^t, r_i^t\}_{t=1}^T$, in environments in which there is only object $o_i$, and a random policy is used to generate actions. We denote these data as 
$\mathcal{D}^\mathrm{single} =\{ 
\{ \mathbf{o}_i^t, \mathbf{a}_i^t, r_i^t
\}_{t=1}^T 
\}_{i\in C_j, \forall j=1, \dots, k, i=1, \dots, m}$. 

We initiate the class template $\mathcal{G}_{C_j}$ for each class $C_j$ randomly, and then use $\mathcal{D}^\mathrm{single}$ to learn it, except the contribution of the latent parameters, which is learned in the next phase.
In particular, we learn the class template graph by maximizing the log-likelihood of $\log p_\lambda\left(\mathbf{o}_i^{t+1}, r_i^{t} \mid  \mathbf{o}_i^{t}, \mathbf{a}_i^t, \mathcal{G}^{C_j}\right)$, where $\lambda = \{\lambda_s, \lambda_r\}$ are the parameters of dynamics and reward models. 
For the implementation, we use Gated Recurrent Units (GRU)~\cite{chung2014empirical} to learn the dynamics and reward models jointly with the class template graphs. At time step $t$ for each object $i$ with class $C_j$, the inputs to the GRU are $\{\mathrm{pa}_{\mathcal{G}_{C_j}} (\mathbf{o}_i^{t+1})\}_{i=1}^m$ and $\{\mathrm{pa}_{\mathcal{G}_{C_j}} ({r}_i^{t})\}_{i=1}^m$, and the GRU outputs $\mathbf{o}_i^{t+1}$ and $r_i^t$. The learning objective of this step is given below, where we maximize the log-likelihoods of dynamics and reward models and regularize the graph to be sparse:
\begin{equation*}
\label{l_1_pred}
    \mathrm{argmax}_{\lambda, \{\mathcal{G}^{C_j}\}_{j=1}^k} \sum_{t=1}^{T} \sum_{i=1}^{m}\sum_{l=1}^{n}\left(  
    \log p_{\lambda}(o_i.s_l^{t+1}, r_i^t \mid \mathrm{pa}_{{\mathcal{G}_{C(i)}}} (o_i.s_l^{t+1}), \mathrm{pa}_{{\mathcal{G}_{C(i)}}} ({r}_i^{t}) \right) -  \sum_{j=1}^{k}  \left\| \mathcal{G}^{C_j}\right\|_1 
\end{equation*}
where $m$ and $k$ indicate the number of single-object environments and object type classes, respectively, and $\mathrm{pa}_{{\mathcal{G}_{C(i)}}}$ denotes the parents of a variable in the template class graph for the class $C(i)$ of object $o_i$. 
After this step, we fix the learned $\{\mathcal{G}^{C_1}, \mathcal{G}^{C_2}, \ldots, \mathcal{G}^{C_k}\}$ with the exception of the edges from the latent parameters $\theta$, which here we assume are disabled and we will learn in the next step.
In later stages, we will reuse the learned reward model $\lambda_r$ and the class template graphs. 

\subsection{Step 2: Interaction learning and latent parameter inference in multi-object environments.}  

In this step, we consider data $\mathcal{D}^{\mathrm{multi}}$ from $N$ multi-object environments with different object configurations in which the objects can have varying latent parameters. Formally, we define
$\mathcal{D}^\mathrm{multi} =\{ 
\{ \mathbf{o}_1^t, \mathbf{a}_1^t, r_1^t
, \mathbf{o}_2^t, \mathbf{a}_2^t, r_2^t
, \dots,  \mathbf{o}_m^t, \mathbf{a}_m^t, r_m^t
\}_{t=1}^T 
\}_{i=1}^N$. 
In each of these environments, we assume the agent can interact only with one object at a time. On these data, we again extract the objects and their attributes from a sequence of images with pre-trained object-centric methods and classify the objects using the object classifier. 
We use these data to learn the interaction pattern graphs $\mathcal{G}_{C_i, C_j}$ for each pair of classes $C_i$ and $C_j$ and the dynamic interaction graph $\mathcal{G}_{\mathrm{inter}}$  by exploiting the previously learned class template graphs.
In particular, we first learn the action binding $\alpha^t$, and at a second stage, we jointly learn the rest of the dynamic interaction graph $\mathcal{G}_{\mathrm{inter}}$, the interaction patterns $\mathcal{G}_{C_i, C_j}$  for each pair of classes, the object-specific latent parameters $o_i.\boldsymbol{\theta}$ and their edges to the other attributes in $\mathcal{G}_{C_j}$.
We describe these two stages in detail in the following.

\subsubsection{Step 2.1: Learning the action binding} \label{sec:step2.1}
Motivated by \cite{biza2022binding}, we learn the dynamic action binding $\mathbf{\alpha}=\{\mathbf{\alpha}^1, \mathbf{\alpha}^2, \ldots, \mathbf{\alpha}^T\}$ using soft attention networks, which are modified by the single-head self-attention module in the Transformer model~\cite{vaswani2017attention}. 
Specifically, we perform non-linear transformations on the states and actions using multi-layer perceptrons (MLPs) to derive the key $\mathbf{k}^t= \left<f_{k}(\mathbf{o}_i^t), f_{k}(\mathbf{o}_2^t), \ldots, f_{k}(\mathbf{o}_m^t)\right>$, query $\mathbf{q}^t= f_{q}\left(\mathbf{a}^t\right)$, and value $\mathbf{v}^t= f_{v}\left(\mathbf{a}^t\right)$, respectively.
We then compute the attention weights $\mathbf{\alpha}^t=\texttt{softmax}\left({(\mathbf{k}_1^t)^\intercal}\mathbf{q}^t, {(\mathbf{k}_2^t)^\intercal}\mathbf{q}^t, \ldots, {(\mathbf{k}_m^t)^\intercal}\mathbf{q}^t\right)$.
We use the learned attention weights ${\alpha}$ as the action binding selector, as it provides an estimation of the binding affinity from the action to the objects at each time step. 
The soft attention mechanism assigns weights from the action to the objects by multiplying the value vector $\mathbf{v}^t$ with the attention weights ${\alpha}^t$, and then embeds the weighted actions into the dynamics of each object. 
We maintain a fixed structure for the class template graphs and focus on learning the action binding selector by updating $f_k$, $f_q$, and $f_v$. 

\subsubsection{Step 2.2: Learning dynamic interaction graph} \label{sec:step2.2}
As noted in Section~\ref{sec2}, the interaction among objects may change over time and usually occurs in a sparse manner. To learn this dynamic graph, we leverage a sequential latent variable model to infer the object interaction graph. Following the neural relational inference (NRI) works~\cite{kipf2018neural, graber2020dynamic}, we use an encoder to generate the latent variables and subsequently sample the interaction graph based on these variables.
Specifically, we use graph neural networks (GNN) as the encoder module, where the nodes represent the states of each object, and the predicted edges denote the temporal interactions between the objects. In line with the dynamic NRI (dNRI)~\cite{graber2020dynamic}, we use recurrent units to model the temporal relations of the interaction graph. We outline the key components of the model in the following and provide a detailed description in App.~\ref{app:algorithm}.

\textbf{Encoder and prior.} During training, at each time step $t$, we use GNN layers to generate hidden embeddings $\mathbf{h}^t=\texttt{GNN}\left(\mathbf{o}^t_1, \mathbf{o}^t_2, \ldots, \mathbf{o}^t_m\right)$. For each object pair $o_i$ and $o_j$, we then obtain $\mathbf{h}^t_{(i,j)}$.
For the encoder, we use a Gated Recurrent Unit (GRU)~\cite{chung2014empirical}  to model the temporal dependency of the interaction graphs. The inputs to the GRU include the future embeddings of the states, which then generate the updated embeddings: $\mathbf{h}_{(i, j), \mathrm{enc}}^t=\texttt{GRU}_{\mathrm{enc}}(\mathbf{h}^t_{(i,j)}, \mathbf{h}^{t+1}_{(i,j), \mathrm{enc}})$. 
For the prior, we also use a GRU, but in this context, we append $\mathbf{h}^t_{(i,j)}$ with the output of the GRU from the previous step $\mathbf{h}_{(i, j), \mathrm{prior}}^{t-1}$ as the input. In specific terms, we have $\mathbf{h}_{(i, j), \mathrm{prior}}^t=\texttt{GRU}_{\mathrm{prior}}({\mathbf{h}^t_{(i,j)}, \mathbf{h}_{(i, j), \mathrm{prior}}^{t-1}})$.
For both the encoder and the prior, we feed the output embedding from the GRUs to an MLP layer to derive the distribution of the latent variables $q_\phi(\mathbf{z}^t \mid \mathbf{s}^{1:T})$, where $\mathbf{s}^{1:T} = \{ \mathbf{o_1}^t, \dots, \mathbf{o_m}^t \}_{t=1}^T$, and prior distribution $p_\phi(\boldsymbol{z}^t \mid \boldsymbol{s}^{1: t}, \boldsymbol{z}^{1: t-1})$, respectively.
We assume that the encoder outputs a Bernoulli distribution for each edge and the graph $\mathcal{G}_{\text{inter}}$ is sampled using the Gumbel-Softmax trick~\cite{jang2017categorical}. 

\textbf{Decoder} We perform one-step prediction to generate $\hat{\mathbf{o}}_i^{t+1}$ for each object $o_i$, predicting the state dynamics with the learned graphs, including $\mathcal{G}_{\mathrm{inter}}$, as well as the interaction pattern graph $\mathcal{G}_{C_i, C_j}$ for two objects with class $C_i$ and $C_j$. At the same time, our goal is also to learn the interaction pattern graph with sparsity regularization. We also learn the latent parameters, $\boldsymbol{\theta}$ at this stage. Specifically, we also incorporate $C_j.\boldsymbol{\theta}$ and the graph from $\mathcal{G}_{C_j.\boldsymbol{\theta}, \mathbf{o}^t_i}$ for each object $o_i$ with class $C_j$ into the dynamics model and use the sparsity regularization for $\mathcal{G}_{C_j.\boldsymbol{\theta}, \mathbf{o}_i}$. 

Therefore, the learning objectives include maximizing the likelihood of the dynamics model, minimizing the KL divergence between $p_\phi$ and $q_\phi$ to estimate $\boldsymbol{\theta}$, and encouraging the sparsity of the interaction pattern graph $\mathcal{G}_{C_i, C_j}$ and the subgraph from latent parameters to states $\mathcal{G}_{C_i.\boldsymbol{\theta}, \mathbf{o}_i}$:
\begin{equation*}
\begin{split}
     & \mathrm{argmax}_{\lambda_s, \phi, v,k,q,\theta,  \mathcal{G}} 
     \sum_{t=1}^{T} \sum_{i=1}^{m}\sum_{l=1}^{n}\ 
    \log p_{\lambda_s}\!\left( o_i.s_l^{t+1} \mid \mathrm{pa}_{{\mathcal{G}^t_{o_i}}}\!\!(o_i.s_l^{t+1}), 
    \{\mathrm{pa}_{G^t_{o_i, o_k}}\!\!(o_i.s_l^{t+1})\}_{o_k \rightarrow o_i \in \mathcal{G}^t_{\mathrm{inter}}}\!\right) \\
    & -  \sum_{j=1}^{k} \sum_{i=1}^{k} \left\| \mathcal{G}_{C_i, C_j} \right\|_1 - \sum_{i=1}^m \sum_{j=1}^k \left\| \mathcal{G}_{C_j.\boldsymbol{\theta}, \mathbf{o}_i} \right\|_1 
     - \sum_{t=2}^T \mathrm{KL} \left(q_\phi\left(\mathbf{z}^t \mid \mathbf{s}^{1:T}\right) \| p_\phi\left(\boldsymbol{z}^t \mid \boldsymbol{s}^{1: t}, \boldsymbol{z}^{1: t-1}\right)\right) \\
\end{split}
\end{equation*}

where $\mathcal{G}$ is the dynamic interaction graph $\mathcal{G}_{\mathrm{inter}}$, interaction pattern graphs $\{\mathcal{G}_{C_i, C_j} \mid i \in \{1,2,\ldots, k\}, j \in \{1,2,\ldots, k\}\}$ and the subgraph from latent parameters to states $\{\mathcal{G}_{C_j.\boldsymbol{\theta}, \mathbf{o}_i} \mid i \in \{1,2,\ldots, m\}, j \in \{1,2,\ldots, k\}\}$. Similarly, to Definition~\ref{def:daft-mdp}, $\mathrm{pa}_{\mathcal{G}^t_{o_i}} (o_{i}.s_l^{t+1})$ indicates the parents of the attribute $s_l$ for object $o_i$ based on the class template graph $\mathcal{G}_{C(i)}$, where $C(i)$ is the class of $o_i$, and where the action binding $\alpha^t$ activates or deactivates any potential connections from $\mathbf{a}^t$. In the second term of the conditioning, we iterate over the objects $o_k$ that are interacting with $o_i$ at time $t$ in the dynamic interaction graph $o_k \to o_i \in \mathcal{G}^t_{\mathrm{inter}}$. For each of these objects $o_k$ we collect the attributes that interact with $o_i.s_l$ in the instance interaction pattern $\mathcal{G}^t_{o_i, o_k}$ based on interaction pattern graph $\mathcal{G}_{C(i), C(k)}$ for the respective classes $C(i)$ and $C(k)$. $\lambda_s$ and $\phi$ indicate the dynamics model and encoder parameters, while $v,k,q$ are the parameters of MLPs for learning the attention models. 
After this, we have learned all the graphs, dynamics, and reward models that estimate DAFT-MDP.

\subsection{Step 3: Policy learning and imagination in multi-object environments.}
In the first two phases, we have learned the template for our world model, which we can now finetune to new multi-object domains by inferring the environment-specific latent parameters $\boldsymbol{\theta}$ and the interaction graph $\mathcal{G}_{\mathrm{inter}}$.
We again consider several multi-object environments with different object configurations in which the objects can have varying latent parameters.
For each environment, we then use the finetuned environment-specific world model to create a set of imagined trajectories. Finally, we can learn a policy $\pi^*(\mathbf{a}^t| \mathbf{s}^t, \boldsymbol{\theta}, \mathcal{G}_{\mathrm{inter}})$ across different environments, based on the real and imagined trajectories.
We can apply policy learning or planning methods using any RL algorithms. To take full advantage of the estimated models, we use MBRL or planning methods such as model predictive control (MPC)~\cite{camacho2013model} or proximal policy optimization (PPO)~\cite{schulman2017proximal} to learn $\pi^*$. 
Detailed information about the domain parameters is provided in App.~\ref{app:algorithm}.

\subsection{Step 4: Adaptation to a new multi-object environment.}
In a new environment, we apply the policy $\pi^*(\mathbf{a}^t| \mathbf{s}^t, \boldsymbol{\theta}, \mathcal{G}_{\mathrm{inter}})$
by inferring latent parameters $\boldsymbol{\theta}$ and dynamic interaction graphs $\mathcal{G}_{\mathrm{inter}}$ based on a few trajectories, without any policy learning.

\section{Related work}\label{sec:rl}
We shortly summarize the related work in this section and provide a more detailed discussion of related work, including the discussion of each method and a comparison based on the method features in App.~\ref{app:morerelwork}.
Recent work in object-oriented and relational RL has incorporated various inductive biases for modeling object relations into both model-based and model-free RL frameworks.
\citet{zhou2022policy} investigate the benefits of deep learning architectures, such as MLP, self-attention, and deep-sets in goal-conditioned RL with factorized object states.
Likewise, \citet{yoon2023investigation} provide a comprehensive empirical evaluation of pre-trained object-oriented models, such as SA\cite{locatello2020object} and SLATE~\cite{singh2022illiterate}, for model and policy learning in multi-object RL. \citet{mambelli2022compositional} use linear relational networks to model object interactions and learn the policy. Another line of research focuses on learning structured representations among objects and their interactions~\cite{watters2019cobra, zadaianchuk2021selfsupervised, zadaianchuk2022self, Kossen2020Structured, chang2023hierarchical, nakanointeraction, biza2022factored, Kipf2020Contrastive}. Most of these approaches aim to learn an object-wise factorization model, either with structured symbolic input or high-dimensional raw pixels as input. NCS~\cite{chang2023hierarchical} and STEDIE~\cite{nakanointeraction} go further by disentangling action/control-relevant or irrelevant features for each object. Unlike these works, we propose a more fine-grained factored world model that considers the structure among all attributes of the objects as well as the dynamic interaction among all objects.

\section{Experimental evaluation}
We consider a diverse set of RL benchmarks and setups, including modified OpenAI Fetch environments~\cite{zhou2022policy, brockman2016openai, lanier2019curiosity} (symbolic inputs), Spriteworld~\cite{watters2019cobra} (visual inputs), and the Block-stacking benchmark~\cite{plappert2018multi}(visual inputs). 
We compare our approach with several baseline models. These include methods using Deep sets, GNN, and Self-Attention as the inductive bias~\cite{zhou2022policy, li2020towards}, such as SRICS~\cite{zadaianchuk2022self}, STOVE~\cite{Kossen2020Structured}, SMORL~\cite{zadaianchuk2021selfsupervised}, NCS~\cite{chang2023hierarchical}, LRN~\cite{mambelli2022compositional}, and COBRA~\cite{watters2019cobra}. To ensure a fair comparison, we modify these baselines, so that methods that originally only support symbolic input can also be adapted to handle image-based input by using visual encoders to obtain the symbolic states, and add an imagination component. We provide the details of these modifications in App.~\ref{app:exp_baseline}. We provide descriptions of all setups, training and testing domains for each benchmark in App.~\ref{app:exp_setup}. Here we describe the most representative results, but we provide the complete results in App.~\ref{app:exp_res}.

\begin{figure}[t]
    \centering
\includegraphics[width=\linewidth]{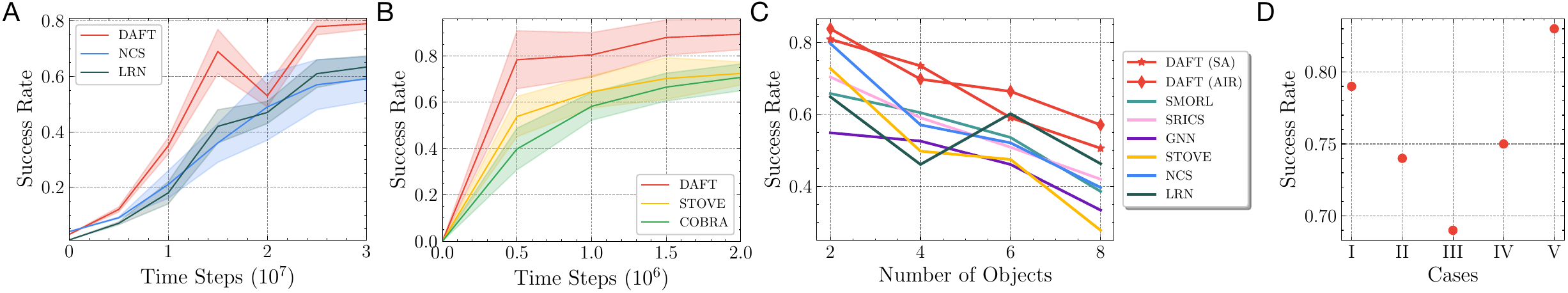}
    \caption{A. The smoothed learning curve for $2$-Push + $2$-Switch (L+S) with different friction coefficients for each object (for clarity, we show only the top three methods in terms of the success rate); B. The smoothed learning curve for the object comparison task in Spriteworld with unseen object numbers, combinations of colors and shapes (for clarity, we show only the top three methods in terms of the success rate); C. Success rate versus number of blocks in the stacking task, where each block has distinct mass; D. Ablation study on the $2$-Push+$2$-Switch task: I. DAFT-RL w/o latent parameters; II. DAFT-RL w/o factored class template graph; III. DAFT-RL w/o dynamic interaction graph; IV. DAFT-RL w/o factored interaction pattern; V. Original DAFT.} 
    \label{fig:exp}
\end{figure}

\paragraph{Symbolic benchmark: OpenAI Fetch - Push and Switch.}
Following \cite{zhou2022policy}, we modify the OpenAI Gym Fetch environment~\cite{brockman2016openai} to create the $N$-push and $N$-switch benchmarks, where $N$ denotes the number of objects. In the $N$-push task, the agent is trained to push all cubes to their corresponding target positions. Similarly, for the $N$-switch task, the agent is required to flip all switches in the environment. In this benchmark, all inputs are symbolic. 
As a sanity check, we show in App.~\ref{app:exp_res} that our method has comparable results with the baselines in the \textit{single-task mode}, in which we train the model estimation and policy learning individually on the $2$-Push, $3$-Push, $2$-Switch, and $3$-Switch tasks.
To evaluate \textit{compositional generalization}, we train on the set of tasks \{$1$-Push, $1$-Switch, $2$-Push, $2$-Switch\}; while during the test phase, we test the model in different settings as shown in Table~\ref{tbl:push-switch-2}. We consider combinations of skills (denoted by $S$), e.g. $2$-Push+$2$-Switch ($S$), which combines the training tasks $2$-Push, $2$-Switch. We also consider changes in the number of objects (denoted by $O$) and skills, e.g. $3$-Push+$3$-Switch ($S$+$O$), which combines the training tasks but also varies the number of objects.
We also test whether our model can achieve efficient transfer during testing when the objects' latent parameters differ from the training samples (denoted by $L$). Generally, during training, we consider the objects with masses and friction coefficients uniformly sampled from a set of values, and during testing, we test the model with two different sets of values. For example, $2$-Switch ($L$) considers this case. Finally, we consider the challenging setting in which we combine all of these changes. As seen in Table~\ref{tbl:push-switch-2}, in most cases DAFT-RL outperforms the baselines, with a bigger gain in the difficult settings. In this setting, we do not compare with COBRA~\cite{watters2019cobra}, since it expects pixel inputs.
As a representative example, we show the smoothed learning curve for $2$-Push+$2$-Switch (L+S) in Fig.~\ref{fig:exp}A, in which we display the top three methods in terms of success rate.


\begin{table}[t]
\centering
\caption{Average success rate over 3 random seeds for Push \& Switch compositional generalization in terms of the combination of skills (S), changing the number of objects (O), and changing latent parameters (L) with respect to training. The numbers in bold highlight the top-performing method.}
\resizebox{\textwidth}{!}{%
\begin{tabular}{@{}llllllllll@{}}
\toprule
\multirow{2}{*}{\textbf{\begin{tabular}[c]{@{}c@{}}Experiment\\ Settings\end{tabular}}} & \multicolumn{9}{c}{\textbf{Method}}                               \\ 
\cmidrule(l){2-10}              & DAFT-RL(Symbolic) & DeepSets  & Self-attention & SRICS & GNN & STOVE & SMORL & NCS & LRN  \\ \midrule
2-Push + 2-Switch (S) &
 $0.881$ \textpm $0.038$ & {$0.498$ \textpm $0.024$} & $0.367$ \textpm $0.027$ & $0.717$ \textpm $0.039$ & $0.633$ \textpm $0.027$ & $0.808$ \textpm $0.032$ & $0.788$ \textpm $0.021$ & \textbf{0.912 \textpm 0.022} & $0.813$ \textpm $0.047$ \\
3-Push + 3-Switch (S+O)
& \textbf{0.805 \textpm 0.024} & {$0.324$ \textpm $0.012$} & $0.183$ \textpm $0.031$ & $0.685$ \textpm $0.019$ & $0.594$ \textpm $0.023$ & $0.672$ \textpm $0.021$ & $0.694$ \textpm $0.017$ & $0.751$ \textpm $0.024$ & $0.715$ \textpm $0.025$\\
2-Push (L)
& \textbf{0.968 \textpm 0.036} & 0.940 \textpm $0.029$ & 0.941 \textpm $0.046$ & 0.922 \textpm $0.042$ & 0.953 \textpm $0.037$ & 0.931 \textpm $0.047$ & 0.964 \textpm $0.023$ & 0.944 \textpm $0.039$ & 0.923 \textpm $0.038$\\ 
2-Switch (L) &
\textbf{0.923 \textpm 0.065} & 0.891 \textpm $0.043$ & 0.912 \textpm $0.054$ & 0.919 \textpm $0.032$ & 0.873 \textpm $0.046$ & 0.824 \textpm $0.027$ & 0.892 \textpm $0.035$ & 0.864 \textpm $0.062$ & 0.907 \textpm $0.048$\\ 
3-Push (L+O) &
\textbf{0.921 \textpm 0.037} & $0.862$ \textpm $0.036$ & $0.907$ \textpm $0.036$ & $0.824$ \textpm $0.036$ & $0.852$ \textpm $0.036$ & $0.915$ \textpm $0.036$ & $0.828$ \textpm $0.036$ & $0.890$ \textpm $0.036$ & $0.886$ \textpm $0.036$\\ 
3-Switch (L+O) &
\textbf{0.903 \textpm 0.023} & $0.451$ \textpm $0.075$ & $0.392$ \textpm $0.045$ & $0.590$ \textpm $0.087$ & $0.744$ \textpm $0.057$ & $0.772$ \textpm $0.079$ & $0.835$ \textpm $0.029$ & $0.818$ \textpm $0.074$ & $0.857$ \textpm $0.032$\\ 
2-Push + 2-Switch (L+S) &
 \textbf{0.793 \textpm 0.026} & $0.384$ \textpm $0.077$ & $0.351$ \textpm $0.032$ & $0.549$ \textpm $0.089$ & $0.476$ \textpm $0.062$ & $0.595$ \textpm $0.043$ & $0.463$ \textpm $0.057$ & $0.591$ \textpm $0.081$ & $0.634$ \textpm $0.045$  \\
3-Push + 3-Switch (L+O+S)
& \textbf{0.783 \textpm 0.025} & $0.256$ \textpm $0.045$ & $0.115$ \textpm $0.029$ & $0.531$ \textpm $0.082$ & $0.362$ \textpm $0.035$ & $0.531$ \textpm $0.042$ & $0.525$ \textpm $0.051$ & $0.529$ \textpm $0.027$ & $0.418$ \textpm $0.092$ \\
\bottomrule
\end{tabular}%
}
\label{tbl:push-switch-2}
\end{table}

\paragraph{Image benchmark: Spriteworld}
We consider four tasks in Spriteworld~\cite{watters2019cobra}: object goal, interaction, comparison, and property comparison. 
Given that the observations are in pixel space, we use pre-trained SA~\cite{locatello2020object} or AIR~\cite{eslami2016attend} as pre-trained encoders to obtain the object factored states. Following~\cite{yoon2023investigation}, we generate the datasets with varying numbers of objects placed randomly for pre-training the object-centric model. 
We consider cases in which we vary the number of objects, have unseen combinations of colors for the objects or unseen combinations of shapes in the target task, and provide the results in Table~\ref{table8}, showing that DAFT-RL with Slot Attention outperforms all baselines, followed closely by DAFT-RL with AIR. In this setting, we do not show the results from DeepSets or Self-Attention, because of bad performance, but we provide the results in App.~\ref{app:exp_res}. We show the smoothed learning curve for the object comparison task in Fig.~\ref{fig:exp}B, in which we only display the top three methods.

\begin{table}[t]
\centering
\caption{Average success rate over 3 random seeds for Spriteworld with unseen object numbers, color and shape combinations. The numbers in bold highlight the top-performing method.}
\resizebox{\textwidth}{!}{%
\begin{tabular}{@{}llllllllll@{}}
\toprule
\multirow{2}{*}{\textbf{\begin{tabular}[c]{@{}c@{}}Experiment\\ Settings\end{tabular}}} & \multicolumn{9}{c}{\textbf{Method}}                               \\ 
\cmidrule(l){2-10}              & DAFT-RL (SA) & DAFT-RL (AIR)  & SMORL & SRICS & GNN & STOVE & COBRA & NCS & LRN  \\ \midrule
Object Goal &
\textbf{0.897 \textpm 0.034} & $0.891$ \textpm $0.039$ & $0.668$ \textpm $0.073$ & $0.710$ \textpm $0.060$ & $0.390$ \textpm $0.101$ & $0.636$ \textpm $0.048$ & $0.663$ \textpm $0.051$ & $0.847$ \textpm $0.022$ & $0.759$ \textpm $0.067$ \\
Object Interaction &
\textbf{0.890 \textpm 0.070} & $0.874$ \textpm $0.055$ & $0.685$ \textpm $0.063$ & $0.742$ \textpm $0.104$ & $0.320$ \textpm $0.138$ & $0.605$ \textpm $0.096$ & $0.668$ \textpm $0.073$ & $0.837$ \textpm $0.062$ & $0.749$ \textpm $0.115$ \\
Object Comparison &
\textbf{0.893 \textpm 0.067} & $0.879$ \textpm $0.061$ & $0.834$ \textpm $0.050$ & $0.735$ \textpm $0.072$ & $0.399$ \textpm $0.121$ & $0.724$ \textpm $0.088$ & $0.707$ \textpm $0.057$ & $0.716$ \textpm $0.054$ & $0.760$ \textpm $0.098$ \\
Property Comparison &
\textbf{0.907 \textpm 0.066} & $0.879$ \textpm $0.073$ & $0.805$ \textpm $0.089$ & $0.732$ \textpm $0.095$ & $0.295$ \textpm $0.169$ & $0.640$ \textpm $0.105$ & $0.660$ \textpm $0.083$ & $0.831$ \textpm $0.112$ & $0.748$ \textpm $0.089$ \\
\bottomrule
\end{tabular}%
}
\label{table8}
\end{table}

\paragraph{Image benchmark: Block-stacking}
In the Block-stacking benchmark~\cite{plappert2018multi}, the task is to stack all the blocks into the target position. We use the pre-trained encoders to obtain the object attributes and MPC~\cite{camacho2013model} to learn the optimal policy. We follow the experimental setup in the multi-step planning task configuration in~\cite{nakanointeraction}, which is specified in App.~\ref{app:exp_setup}. 
As a sanity check, we show in App.~\ref{app:exp_res} that DAFT-RL provides comparable performances to the baselines in the \emph{single-task mode}. We train on a varying number of objects and objects with different masses and then use the trained model to test on the domains where there exist an unseen number of objects and unseen combinations of masses. 
We show the results in Table~\ref{tbl:stacking-1} and Fig.~\ref{fig:exp}C for varying numbers of blocks. Our approaches consistently outperform baselines, with bigger gains for more objects.

\begin{table}[t]
\centering
\caption{
Average success rate over 3 random seeds for Block-stacking with unseen object numbers and mass combinations. The numbers in bold highlight the top-performing method.}
\resizebox{\textwidth}{!}{%
\begin{tabular}{@{}lllllllll@{}}
\toprule
\multirow{2}{*}{\textbf{\begin{tabular}[c]{@{}c@{}}Experiment\\ Settings\end{tabular}}} & \multicolumn{8}{c}{\textbf{Method}}                               \\ 
\cmidrule(l){2-9}             & DAFT-RL (SA) & DAFT-RL (AIR)  & SMORL & SRICS & GNN & STOVE  & NCS & LRN  \\ \midrule
2 Blocks 
&  $0.809$ \textpm $0.019$ & \textbf{0.838 \textpm 0.030} & $0.658$ \textpm $0.028$ & $0.704$ \textpm $0.016$ & $0.549$ \textpm $0.016$ & $0.728$ \textpm $0.044$ & $0.797$ \textpm $0.035$ & $0.649$ \textpm $0.026$ \\ 
4 Blocks &
\textbf{0.735 \textpm 0.032} & $0.698$ \textpm $0.022$ & $0.605$ \textpm $0.020$ & $0.591$ \textpm $0.049$ & $0.526$ \textpm $0.041$ & $0.498$ \textpm $0.013$ & $0.571$ \textpm $0.026$ & $0.461$ \textpm $0.028$ \\ 
6 blocks &
$0.591$ \textpm $0.025$ & \textbf{0.664 \textpm 0.017} & $0.536$ \textpm $0.040$ & $0.509$ \textpm $0.043$ & $0.461$ \textpm $0.088$ & $0.475$ \textpm $0.023$ & $0.521$ \textpm $0.049$ & $0.602$ \textpm $0.097$ \\ 
8 blocks &
$0.506$ \textpm $0.083$ & \textbf{0.571 \textpm 0.039} & $0.386$ \textpm $0.062$ & $0.420$ \textpm $0.061$ & $0.334$ \textpm $0.047$ & $0.278$ \textpm $0.086$ & $0.397$ \textpm $0.052$ & $0.463$ \textpm $0.077$ \\ 
\bottomrule
\end{tabular}%
}
\label{tbl:stacking-1}
\end{table}

\paragraph{Ablation studies}
To evaluate the effectiveness of each component in DAFT, we conduct the following ablation studies: a) DAFT w/o latent parameters,
b) DAFT w/o class template graphs, c) DAFT w/o dynamic interaction graph, and d) DAFT w/o interaction pattern graphs. As a representative result, we show the results of $2$-Push+$2$-Switch (L+S) in Fig.~\ref{fig:exp}D. The results further indicate the contribution of each part to the effectiveness of final policy learning. 
We also provide visualizations of the learned graphs in App.~\ref{app:exp_res}, which show that our model is capable of learning the true causal of a single object in symbolic cases. For pixel inputs, we do not have the ground-truth causal graphs.

\section{Conclusions and Future work}

We proposed Dynamic Attribute FacTored RL (DAFT-RL), a framework that leverages learned attribute-factored representations with dynamic graphs. For each class of object, we learned a class template graph that describes how the dynamics and reward of a object of this class factorizes according to its attributes, as well as an interaction pattern graph that describes how it interacts with objects of different classes at the attribute level. We also learned interactions between objects and with the agent with a dynamic graph. Through this template world model
we learned a policy that can then be directly applied in a new environment by just estimating the interactions and latent parameters. We showed that DAFT-RL  outperforms the state-of-the-art in three compositional generalization benchmarks. In future work, we plan to investigate jointly learning the representations from pixels, the class template graphs, interaction pattern graph, dynamic interation graphs and the policy.

\clearpage
\bibliographystyle{unsrtnat}
\bibliography{ref}

\begin{thebibliography}{48}
\providecommand{\natexlab}[1]{#1}
\providecommand{\url}[1]{\texttt{#1}}
\expandafter\ifx\csname urlstyle\endcsname\relax
  \providecommand{\doi}[1]{doi: #1}\else
  \providecommand{\doi}{doi: \begingroup \urlstyle{rm}\Url}\fi

\bibitem[Hafner et~al.(2019)Hafner, Lillicrap, Fischer, Villegas, Ha, Lee, and
  Davidson]{hafner2019learning}
Danijar Hafner, Timothy Lillicrap, Ian Fischer, Ruben Villegas, David Ha,
  Honglak Lee, and James Davidson.
\newblock Learning latent dynamics for planning from pixels.
\newblock In \emph{International Conference on Machine Learning (ICML)}, pages
  2555--2565. PMLR, 2019.

\bibitem[Hafner et~al.(2020)Hafner, Lillicrap, Ba, and Norouzi]{hafnerdream}
Danijar Hafner, Timothy Lillicrap, Jimmy Ba, and Mohammad Norouzi.
\newblock Dream to control: Learning behaviors by latent imagination.
\newblock In \emph{International Conference on Learning Representations
  (ICLR)}, 2020.
\newblock URL \url{https://openreview.net/forum?id=S1lOTC4tDS}.

\bibitem[Micheli et~al.(2023)Micheli, Alonso, and
  Fleuret]{micheli2022transformers}
Vincent Micheli, Eloi Alonso, and Fran{\c{c}}ois Fleuret.
\newblock Transformers are sample efficient world models.
\newblock In \emph{International conference on Learning Representations
  (ICLR)}, 2023.

\bibitem[Li et~al.(2020)Li, Jabri, Darrell, and Agrawal]{li2020towards}
Richard Li, Allan Jabri, Trevor Darrell, and Pulkit Agrawal.
\newblock Towards practical multi-object manipulation using relational
  reinforcement learning.
\newblock In \emph{2020 IEEE international conference on robotics and
  automation (ICRA)}, pages 4051--4058. IEEE, 2020.

\bibitem[Zadaianchuk et~al.(2021)Zadaianchuk, Seitzer, and
  Martius]{zadaianchuk2021selfsupervised}
Andrii Zadaianchuk, Maximilian Seitzer, and Georg Martius.
\newblock Self-supervised visual reinforcement learning with object-centric
  representations.
\newblock In \emph{International Conference on Learning Representations
  (ICLR)}, 2021.
\newblock URL \url{https://openreview.net/forum?id=xppLmXCbOw1}.

\bibitem[Zadaianchuk et~al.(2022)Zadaianchuk, Martius, and
  Yang]{zadaianchuk2022self}
Andrii Zadaianchuk, Georg Martius, and Fanny Yang.
\newblock Self-supervised reinforcement learning with independently
  controllable subgoals.
\newblock In \emph{Conference on Robot Learning (CoRL)}, pages 384--394. PMLR,
  2022.
\newblock URL \url{https://openreview.net/forum?id=TEQWRlncJVm}.

\bibitem[Mambelli et~al.(2022)Mambelli, Tr{\"a}uble, Bauer, Sch{\"o}lkopf, and
  Locatello]{mambelli2022compositional}
Davide Mambelli, Frederik Tr{\"a}uble, Stefan Bauer, Bernhard Sch{\"o}lkopf,
  and Francesco Locatello.
\newblock Compositional multi-object reinforcement learning with linear
  relation networks.
\newblock In \emph{ICLR2022 Workshop on the Elements of Reasoning: Objects,
  Structure and Causality}, 2022.
\newblock URL \url{https://openreview.net/forum?id=HFUxPr_I5ec}.

\bibitem[Kossen et~al.(2020)Kossen, Stelzner, Hussing, Voelcker, and
  Kersting]{Kossen2020Structured}
Jannik Kossen, Karl Stelzner, Marcel Hussing, Claas Voelcker, and Kristian
  Kersting.
\newblock Structured object-aware physics prediction for video modeling and
  planning.
\newblock In \emph{International Conference on Learning Representations
  (ICLR)}, 2020.
\newblock URL \url{https://openreview.net/forum?id=B1e-kxSKDH}.

\bibitem[Watters et~al.(2019)Watters, Matthey, Bosnjak, Burgess, and
  Lerchner]{watters2019cobra}
Nicholas Watters, Loic Matthey, Matko Bosnjak, Christopher~P Burgess, and
  Alexander Lerchner.
\newblock Cobra: Data-efficient model-based rl through unsupervised object
  discovery and curiosity-driven exploration.
\newblock \emph{arXiv preprint arXiv:1905.09275}, 2019.

\bibitem[Goyal et~al.(2021)Goyal, Lamb, Hoffmann, Sodhani, Levine, Bengio, and
  Sch{\"o}lkopf]{goyal2021recurrent}
Anirudh Goyal, Alex Lamb, Jordan Hoffmann, Shagun Sodhani, Sergey Levine,
  Yoshua Bengio, and Bernhard Sch{\"o}lkopf.
\newblock Recurrent independent mechanisms.
\newblock In \emph{International Conference on Learning Representations
  (ICLR)}, 2021.
\newblock URL \url{https://openreview.net/forum?id=mLcmdlEUxy-}.

\bibitem[Choi et~al.(2023)Choi, Lee, Wang, Sohn, and Lee]{choi2023unsupervised}
Jongwook Choi, Sungtae Lee, Xinyu Wang, Sungryull Sohn, and Honglak Lee.
\newblock Unsupervised object interaction learning with counterfactual dynamics
  models.
\newblock In \emph{Workshop on Reincarnating Reinforcement Learning at ICLR
  2023}, 2023.
\newblock URL \url{https://openreview.net/forum?id=dYjH8Nv81K}.

\bibitem[Chang et~al.(2023)Chang, Dayan, Meier, Griffiths, Levine, and
  Zhang]{chang2023hierarchical}
Michael Chang, Alyssa~Li Dayan, Franziska Meier, Thomas~L. Griffiths, Sergey
  Levine, and Amy Zhang.
\newblock Hierarchical abstraction for combinatorial generalization in object
  rearrangement.
\newblock In \emph{The Eleventh International Conference on Learning
  Representations (ICLR)}, 2023.
\newblock URL \url{https://openreview.net/forum?id=fGG6vHp3W9W}.

\bibitem[Yoon et~al.(2023)Yoon, Wu, Bae, and Ahn]{yoon2023investigation}
Jaesik Yoon, Yi-Fu Wu, Heechul Bae, and Sungjin Ahn.
\newblock An investigation into pre-training object-centric representations for
  reinforcement learning.
\newblock \emph{International Conference on Machine Learning (ICML)}, 2023.

\bibitem[Zhao et~al.(2022)Zhao, Kong, Walters, and Wong]{zhao2022toward}
Linfeng Zhao, Lingzhi Kong, Robin Walters, and Lawson~LS Wong.
\newblock Toward compositional generalization in object-oriented world
  modeling.
\newblock In \emph{International Conference on Machine Learning (ICML)}, pages
  26841--26864. PMLR, 2022.

\bibitem[Zhou et~al.(2022)Zhou, Kumar, Finn, and Rajeswaran]{zhou2022policy}
Allan Zhou, Vikash Kumar, Chelsea Finn, and Aravind Rajeswaran.
\newblock Policy architectures for compositional generalization in control.
\newblock In \emph{ICML Workshop on Spurious Correlations, Invariance and
  Stability}, 2022.

\bibitem[Biza et~al.(2022{\natexlab{a}})Biza, Kipf, Klee, Platt, van~de Meent,
  and Wong]{biza2022factored}
Ondrej Biza, Thomas Kipf, David Klee, Robert Platt, Jan-Willem van~de Meent,
  and Lawson~LS Wong.
\newblock Factored world models for zero-shot generalization in robotic
  manipulation.
\newblock \emph{arXiv preprint arXiv:2202.05333}, 2022{\natexlab{a}}.

\bibitem[Nakano et~al.(2023)Nakano, Suzuki, and Matsuo]{nakanointeraction}
Akihiro Nakano, Masahiro Suzuki, and Yutaka Matsuo.
\newblock Interaction-based disentanglement of entities for object-centric
  world models.
\newblock In \emph{The Eleventh International Conference on Learning
  Representations (ICLR)}, 2023.

\bibitem[Sallans and Hinton(2004)]{sallans2004reinforcement}
Brian Sallans and Geoffrey~E Hinton.
\newblock Reinforcement learning with factored states and actions.
\newblock \emph{The Journal of Machine Learning Research}, 5:\penalty0
  1063--1088, 2004.

\bibitem[Kearns and Koller(1999)]{kearns1999efficient}
Michael Kearns and Daphne Koller.
\newblock Efficient reinforcement learning in factored mdps.
\newblock In \emph{Proceedings of the 16th international joint conference on
  Artificial Intelligence (IJCAI)}, pages 740--747, 1999.

\bibitem[Boutilier et~al.(2000)Boutilier, Dearden, and
  Goldszmidt]{boutilier2000stochastic}
Craig Boutilier, Richard Dearden, and Mois{\'e}s Goldszmidt.
\newblock Stochastic dynamic programming with factored representations.
\newblock \emph{Artificial intelligence}, 121\penalty0 (1-2):\penalty0 49--107,
  2000.

\bibitem[Guestrin et~al.(2003{\natexlab{a}})Guestrin, Koller, Parr, and
  Venkataraman]{guestrin2003efficient}
Carlos Guestrin, Daphne Koller, Ronald Parr, and Shobha Venkataraman.
\newblock Efficient solution algorithms for factored mdps.
\newblock \emph{Journal of Artificial Intelligence Research}, 19:\penalty0
  399--468, 2003{\natexlab{a}}.

\bibitem[Hallak et~al.(2015)Hallak, Schnitzler, Mann, and
  Mannor]{hallak2015off}
Assaf Hallak, Fran{\c{c}}ois Schnitzler, Timothy Mann, and Shie Mannor.
\newblock Off-policy model-based learning under unknown factored dynamics.
\newblock In \emph{International Conference on Machine Learning (ICML)}, pages
  711--719. PMLR, 2015.

\bibitem[Katt et~al.(2019)Katt, Oliehoek, and Amato]{factoredPOMDP}
Sammie Katt, Frans~A. Oliehoek, and Christopher Amato.
\newblock Bayesian reinforcement learning in factored pomdps.
\newblock In \emph{Proceedings of the 18th International Conference on
  Autonomous Agents and MultiAgent Systems (AAMAS)}, AAMAS '19, page 7–15,
  Richland, SC, 2019.
\newblock ISBN 9781450363099.

\bibitem[Gardiol and Kaelbling(2003)]{gardiol2003envelope}
Natalia Gardiol and Leslie Kaelbling.
\newblock Envelope-based planning in relational mdps.
\newblock \emph{Advances in Neural Information Processing Systems (NIPS)}, 16,
  2003.

\bibitem[Guestrin et~al.(2003{\natexlab{b}})Guestrin, Koller, Gearhart, and
  Kanodia]{guestrin2003generalizing}
Carlos Guestrin, Daphne Koller, Chris Gearhart, and Neal Kanodia.
\newblock Generalizing plans to new environments in relational mdps.
\newblock In \emph{Proceedings of the 18th International Joint Conference on
  Artificial Intelligence (IJCAI)}, pages 1003--1010, 2003{\natexlab{b}}.

\bibitem[Van~Otterlo(2005)]{van2005survey}
Martijn Van~Otterlo.
\newblock A survey of reinforcement learning in relational domains.
\newblock \emph{Centre for Telematics and Information Technology (CTIT)
  University of Twente, Tech. Rep}, 2005.

\bibitem[Diuk et~al.(2008)Diuk, Cohen, and Littman]{diuk2008object}
Carlos Diuk, Andre Cohen, and Michael~L Littman.
\newblock An object-oriented representation for efficient reinforcement
  learning.
\newblock In \emph{Proceedings of the 25th International Conference on Machine
  Learning (ICML)}, pages 240--247, 2008.

\bibitem[Scholz et~al.(2014)Scholz, Levihn, Isbell, and
  Wingate]{pmlr-v32-scholz14}
Jonathan Scholz, Martin Levihn, Charles Isbell, and David Wingate.
\newblock A physics-based model prior for object-oriented mdps.
\newblock In \emph{International Conference on Machine Learning (ICML)}, pages
  1089--1097. PMLR, 2014.

\bibitem[Wandzel et~al.(2019)Wandzel, Oh, Fishman, Kumar, Wong, and
  Tellex]{OO-POMDP}
Arthur Wandzel, Yoonseon Oh, Michael Fishman, Nishanth Kumar, Lawson~L.S. Wong,
  and Stefanie Tellex.
\newblock Multi-object search using object-oriented pomdps.
\newblock In \emph{2019 International Conference on Robotics and Automation
  (ICRA)}, pages 7194--7200, 2019.
\newblock \doi{10.1109/ICRA.2019.8793888}.

\bibitem[Locatello et~al.(2020)Locatello, Weissenborn, Unterthiner, Mahendran,
  Heigold, Uszkoreit, Dosovitskiy, and Kipf]{locatello2020object}
Francesco Locatello, Dirk Weissenborn, Thomas Unterthiner, Aravindh Mahendran,
  Georg Heigold, Jakob Uszkoreit, Alexey Dosovitskiy, and Thomas Kipf.
\newblock Object-centric learning with slot attention.
\newblock \emph{Advances in Neural Information Processing Systems (NeurIPS)},
  33:\penalty0 11525--11538, 2020.

\bibitem[Eslami et~al.(2016)Eslami, Heess, Weber, Tassa, Szepesvari, Hinton,
  et~al.]{eslami2016attend}
SM~Eslami, Nicolas Heess, Theophane Weber, Yuval Tassa, David Szepesvari,
  Geoffrey~E Hinton, et~al.
\newblock Attend, infer, repeat: Fast scene understanding with generative
  models.
\newblock \emph{Advances in neural information processing systems}, 29, 2016.

\bibitem[Huang et~al.(2022)Huang, Feng, Lu, Magliacane, and
  Zhang]{huang2021adarl}
Biwei Huang, Fan Feng, Chaochao Lu, Sara Magliacane, and Kun Zhang.
\newblock Ada{RL}: What, where, and how to adapt in transfer reinforcement
  learning.
\newblock In \emph{International Conference on Learning Representations
  (ICLR)}, 2022.
\newblock URL \url{https://openreview.net/forum?id=8H5bpVwvt5}.

\bibitem[Feng et~al.(2022)Feng, Huang, Zhang, and Magliacane]{feng2022factored}
Fan Feng, Biwei Huang, Kun Zhang, and Sara Magliacane.
\newblock Factored adaptation for non-stationary reinforcement learning.
\newblock In Alice~H. Oh, Alekh Agarwal, Danielle Belgrave, and Kyunghyun Cho,
  editors, \emph{Advances in Neural Information Processing Systems (NeurIPS)},
  2022.

\bibitem[Vaswani et~al.(2017)Vaswani, Shazeer, Parmar, Uszkoreit, Jones, Gomez,
  Kaiser, and Polosukhin]{vaswani2017attention}
Ashish Vaswani, Noam Shazeer, Niki Parmar, Jakob Uszkoreit, Llion Jones,
  Aidan~N Gomez, {\L}ukasz Kaiser, and Illia Polosukhin.
\newblock Attention is all you need.
\newblock \emph{Advances in neural information processing systems (NIPS)}, 30,
  2017.

\bibitem[Biza et~al.(2022{\natexlab{b}})Biza, Platt, van~de Meent, Wong, and
  Kipf]{biza2022binding}
Ondrej Biza, Robert Platt, Jan-Willem van~de Meent, Lawson~L.S. Wong, and
  Thomas Kipf.
\newblock Binding actions to objects in world models.
\newblock In \emph{ICLR2022 Workshop on the Elements of Reasoning: Objects,
  Structure and Causality}, 2022{\natexlab{b}}.
\newblock URL \url{https://openreview.net/forum?id=HImz8BuUclc}.

\bibitem[Kipf et~al.(2018)Kipf, Fetaya, Wang, Welling, and
  Zemel]{kipf2018neural}
Thomas Kipf, Ethan Fetaya, Kuan-Chieh Wang, Max Welling, and Richard Zemel.
\newblock Neural relational inference for interacting systems.
\newblock In \emph{International Conference on Machine Learning (ICML)}, pages
  2688--2697. PMLR, 2018.

\bibitem[Graber and Schwing(2020)]{graber2020dynamic}
Colin Graber and Alexander~G Schwing.
\newblock Dynamic neural relational inference.
\newblock In \emph{Proceedings of the IEEE/CVF Conference on Computer Vision
  and Pattern Recognition (CVPR)}, pages 8513--8522, 2020.

\bibitem[Murphy(2002)]{Murphy_DBN}
Kevin Murphy.
\newblock Dynamic bayesian networks: Representation, inference and learning.
\newblock \emph{UC Berkeley, Computer Science Division}, 2002.

\bibitem[Chung et~al.(2014)Chung, Gulcehre, Cho, and
  Bengio]{chung2014empirical}
Junyoung Chung, Caglar Gulcehre, KyungHyun Cho, and Yoshua Bengio.
\newblock Empirical evaluation of gated recurrent neural networks on sequence
  modeling.
\newblock \emph{NIPS 2014 Deep Learning and Representation Learning Workshop},
  2014.

\bibitem[Jang et~al.(2017)Jang, Gu, and Poole]{jang2017categorical}
Eric Jang, Shixiang Gu, and Ben Poole.
\newblock Categorical reparameterization with gumbel-softmax.
\newblock In \emph{International Conference on Learning Representations
  (ICLR)}, 2017.
\newblock URL \url{https://openreview.net/forum?id=rkE3y85ee}.

\bibitem[Camacho and Alba(2013)]{camacho2013model}
Eduardo~F Camacho and Carlos~Bordons Alba.
\newblock \emph{Model predictive control}.
\newblock Springer science \& business media, 2013.

\bibitem[Schulman et~al.(2017)Schulman, Wolski, Dhariwal, Radford, and
  Klimov]{schulman2017proximal}
John Schulman, Filip Wolski, Prafulla Dhariwal, Alec Radford, and Oleg Klimov.
\newblock Proximal policy optimization algorithms.
\newblock \emph{arXiv preprint arXiv:1707.06347}, 2017.

\bibitem[Singh et~al.(2022)Singh, Deng, and Ahn]{singh2022illiterate}
Gautam Singh, Fei Deng, and Sungjin Ahn.
\newblock Illiterate {DALL}-e learns to compose.
\newblock In \emph{International Conference on Learning Representations
  (ICML)}, 2022.
\newblock URL \url{https://openreview.net/forum?id=h0OYV0We3oh}.

\bibitem[Kipf et~al.(2020)Kipf, van~der Pol, and Welling]{Kipf2020Contrastive}
Thomas Kipf, Elise van~der Pol, and Max Welling.
\newblock Contrastive learning of structured world models.
\newblock In \emph{International Conference on Learning Representations
  (ICLR)}, 2020.
\newblock URL \url{https://openreview.net/forum?id=H1gax6VtDB}.

\bibitem[Brockman et~al.(2016)Brockman, Cheung, Pettersson, Schneider,
  Schulman, Tang, and Zaremba]{brockman2016openai}
Greg Brockman, Vicki Cheung, Ludwig Pettersson, Jonas Schneider, John Schulman,
  Jie Tang, and Wojciech Zaremba.
\newblock Openai gym.
\newblock \emph{arXiv preprint arXiv:1606.01540}, 2016.

\bibitem[Lanier(2019)]{lanier2019curiosity}
John~Banister Lanier.
\newblock \emph{Curiosity-driven multi-criteria hindsight experience replay}.
\newblock University of California, Irvine, 2019.

\bibitem[Plappert et~al.(2018)Plappert, Andrychowicz, Ray, McGrew, Baker,
  Powell, Schneider, Tobin, Chociej, Welinder, et~al.]{plappert2018multi}
Matthias Plappert, Marcin Andrychowicz, Alex Ray, Bob McGrew, Bowen Baker,
  Glenn Powell, Jonas Schneider, Josh Tobin, Maciek Chociej, Peter Welinder,
  et~al.
\newblock Multi-goal reinforcement learning: Challenging robotics environments
  and request for research.
\newblock \emph{arXiv preprint arXiv:1802.09464}, 2018.

\bibitem[Veerapaneni et~al.(2020)Veerapaneni, Co-Reyes, Chang, Janner, Finn,
  Wu, Tenenbaum, and Levine]{veerapaneni2020entity}
Rishi Veerapaneni, John~D Co-Reyes, Michael Chang, Michael Janner, Chelsea
  Finn, Jiajun Wu, Joshua Tenenbaum, and Sergey Levine.
\newblock Entity abstraction in visual model-based reinforcement learning.
\newblock In \emph{Conference on Robot Learning (CoRL)}, pages 1439--1456.
  PMLR, 2020.

\end{thebibliography}

   \newpage
    \appendix
    \hrule height 3pt
    \vskip 5mm
    \begin{center}
        \Large{\textbf{Appendix for ``Learning dynamic attribute-factored world models for
efficient multi-object reinforcement learning''}}\\
    \end{center}
    \vskip 3mm
    \hrule height 1pt
    \setcounter{figure}{0}
    \renewcommand{\thefigure}{A\arabic{figure}}
    \setcounter{table}{0}
    \renewcommand{\thetable}{A\arabic{table}}
    \setcounter{algocf}{0}
    \renewcommand{\thealgocf}{A\arabic{algocf}}
    \renewcommand{\thesection}{\Alph{section}}
    \renewcommand{\theequation}{A\arabic{equation}}
    \setcounter{equation}{0}
\appendix

\section{Detailed discussion and comparison with related work}
\label{app:morerelwork}

In this section we discuss the related work. We first shortly discuss related work in terms of our modelling assumptions and then provide a comparison with other object-centric RL methods for compositional generalization.

\subsection{Factored, relational and object-oriented MDPs}
Our modelling assumptions, formalized as a Dynamic Attribute FacTored Markov Decision Process (DAFT-MDP) in Definition~\ref{def:daft-mdp}, are related to the literature on factored and object-oriented (PO)MDPs. In particular DAFT-MDPs are an first order extension with class template graphs, interaction patterns and interaction graphs of factored (PO)MDPs \cite{sallans2004reinforcement,kearns1999efficient,boutilier2000stochastic,guestrin2003efficient,hallak2015off, factoredPOMDP}. In particular they define a family of factored POMDPs, in which the objects $\mathcal{O}$ can vary across environments in numbers, types and latent parameters. 
This extension takes inspiration from Relational MDPs \cite{gardiol2003envelope,guestrin2003generalizing,van2005survey} and their literature, especially Object-Oriented (PO)MDPs \cite{diuk2008object,pmlr-v32-scholz14, OO-POMDP}, which also define a family of MDPs that vary in terms of objects and types. Most of these methods are focusing on discrete states and often define relations or states as first order predicates, with the exception of \cite{pmlr-v32-scholz14}, which propose a physics-based approach that circumvents these issues, allowing for more realistic robotic applications.

As opposed to these works, we also consider continuous state spaces and propose a more fine-grained factorization of the transition and reward at the object \emph{attribute} level, based on estimating class template graphs  and interaction pattern graphs. Additionally, we consider object-specific latent parameters that modulate the dynamics of each object, and use \emph{dynamic} graphs to account for the sparsity of interactions between objects and between the agent and an object.

\subsection{Compositional generalization with object-centric RL approaches}

Table~\ref{tab:prior} provides a comparison of the object-centric RL methods discussed in Section~\ref{sec:rl}. The criteria for comparison include the extent of factorization in the dynamics, the interaction modeling, and the model's ability to adapt to changes in the environment with changing latent factors.

\begin{table}[h]
\centering
\begin{minipage}[b]{0.9\linewidth}
\centering
\caption{Comparison of different object-oriented RL methods for compositional generalization. A check denotes that a method has an attribute, whereas a cross denotes the opposite.
\label{tab:prior} }
\resizebox{\textwidth}{!}{
\begin{tabular}{lccc}
\toprule
{\bf Methods} &
  \begin{tabular}[c]{@{}c@{}}\bf Factored  \\ \bf Dynamics\end{tabular} &
  \begin{tabular}[c]{@{}c@{}}\bf  Interaction  \\ \bf Modeling \end{tabular} &
  \begin{tabular}[c]{@{}c@{}}\bf  Generalize to  \\ \bf Changing Latent Factors \end{tabular}  
  \\ \hline
{GNN~\cite{li2020towards}} & Object & Fully-connected & \xmark    \\
{Self-attention~\cite{zhou2022policy}} & Object & Fully-connected & \xmark    \\
{Deep Sets~\cite{zhou2022policy}} & Object & Fully-connected & \xmark    \\
{LRN~\cite{mambelli2022compositional}}  & Object  & Fully-connected & \xmark   \\
{COBRA~\cite{watters2019cobra}} & Object & \xmark & \xmark    \\
{FWM~\cite{biza2022factored}} & Object & Fully-connected & \xmark    \\
{SMORL~\cite{zadaianchuk2021selfsupervised}}  & Object  & \xmark &  \xmark   \\
{STOVE~\cite{Kossen2020Structured}} & Object \& pre-determined attributes \footnote{ {\tiny STOVE focuses specifically on these attributes: (position, velocity, size, others)}.}  & Fully-connected & \xmark   \\
{SRICS~\cite{zadaianchuk2022self}} & Object  & Sparse \& dynamic (object-level) &   \xmark \\
{NCS~\cite{chang2023hierarchical}} & Object \& action/dynamics-relevant & Fully-connected  & \xmark     \\
{STEDIE~\cite{nakanointeraction}} & Object \& interaction-relevant\footnote{ {\tiny STEDIE learns which attributes interact during object interactions, similar to our interaction pattern graphs.}} & Fully-connected (object-level), factored (interaction-relevant) & \xmark \\
\rowcolor{ourmethod} {DAFT-RL} & Object \& attribute  & Sparse, dynamic \& factored  (object-level \& attribute-level) & \cmark     \\
\bottomrule
\end{tabular}
}
\end{minipage}
\end{table}
In the context of factored dynamics, most of these related works take into consideration the object-factored states. Approaches such as STEDIE~\cite{nakanointeraction}, NCS~\cite{chang2023hierarchical}, and STOVE~\cite{Kossen2020Structured} go a step further to disentangle the states related to interaction, dynamics, or actions from other state variables, such as color, as mentioned in the Table. However, our model aims to offer a more comprehensive approach to factored dynamics, providing an attribute-level factorization for each object. For example this can include factorizing the transition function of dynamics-related factors in terms of position, velocity and mass, and  factorizing action-irrelevant factors like color and shape.
In terms of interaction modeling, many approaches assume fully-connected interaction patterns, where interactions occur between every pair of objects within a given scene. However, in many real-world scenarios, this is often inaccurate as the object interaction usually happens in a sparse and dynamic manner. An exception is SRICS~\cite{zadaianchuk2022self}, which takes into account dynamic interaction modeling. This model emphasizes the sparsity of interactions among objects in a scene, with a dynamically changing structure at the object level.
In contrast, STEDIE~\cite{nakanointeraction} employs learned factored dynamics for each object to model interactions, assuming  that only interaction-irrelevant objects will have an impact on others. Nonetheless, from the perspective of object level, the pattern of interaction in STEDIE is still fully-connected.
Our method diverges from these approaches by integrating both dynamic and sparse interaction patterns at the object level, and factored interactions at the attribute level between pairs of objects. We provide a more fine-grained and realistic model of interactions as from the object level, the interaction pattern is dynamic and sparse, and for each object pair, the interaction only happens among some specific attributes. 
Finally, DAFT-RL is the only framework that considers the latent changing factors in the environment, which is also a realistic setting where the agent sometimes cannot observe all essential attribute variables in the system but these unobserved factors do have effects on the dynamics or reward.

\section{Full example with summary of notation}\label{app:fullexample}
\begin{table}[t]
	\centering
	\begin{threeparttable}
		\caption{Summary of notation in this paper. \label{app-table:notation}}
		\begin{tabular}{c|l}
			\toprule 
			Notation						&	Definition	\\
			\hline 
					
				$\mathcal{C} = \{ \mathcal{C}_1,\dots, \mathcal{C}_k\}$			&	Set of object classes\\  
				$\{C_j.s_1, C_j.s_2, \ldots, C_j.s_n\}$			&	Class attributes for class $C_j$ \\	$C_j.\boldsymbol{\theta}$ &  Latent  constant parameters for class $C_j$\\
                $\mathcal{O} = \{ o_1,\dots, o_m\}$
                    & Set of objects in a domain \\
                    $C(i)$       & Class of object $o_i$, in other words $C_j$ such that $o_i \in C_j$\\
				$\{o_i.s_1, o_i.s_2, \ldots, o_i.s_n\}$			&	Object attributes for object $o_i$ with semantics  defined by the class $C(i)$ \\ $o_i.\boldsymbol{\theta}$		&	Latent constant parameters of object $i$ \\
                \hline 
                $\mathbf{x}^t$			& Pixel observation at time step $t$	\\ $\mathbf{x}_i^t$			& Pixel observation of object-$i$ at time step $t$	\\
			$\mathbf{o}_i^t = \{o_i.s_1^t, o_i.s_2^t, \ldots, o_i.s_n^t\}$			&	Symbolic state of object $o_i$ at time step $t$ \\ $\mathbf{s}^t = \{\mathbf{o}_1^t, \dots, \mathbf{o}_m^t\}$			&	Observable state at time step $t$	for environment with $\mathcal{}$ objects\\  $\boldsymbol{\theta} = \{o_1.\boldsymbol{\theta}, \dots, o_m.\boldsymbol{\theta}\}$			&	Latent constant parameters	for environment with $\mathcal{O}$ objects \\ $\mathbf{a}^t_i$			&	Action on object $o_i$  at time step $t$	\\$\mathbf{a}^t = \{\mathbf{a}_1^t, \dots, \mathbf{a}_m^t\}$			&	Action at time step $t$ for environment with $\mathcal{O}$ objects	\\$r_i^t$			&	Reward on object $o_i$ at time step $t$\\ $r^t = \Sigma_i r_i^t$			&	Reward at time step $t$	\\
                    \hline
                    $\mathcal{G}_{C_j}$                          &  Class template graph for object with class $C_j$\\
                    $\mathcal{G}_{C_i, C_j}$                          & Interaction pattern graph for objects with  class $C_i$ and $C_j$ \\
                    $\boldsymbol{\alpha}^t$			&	Action selector at time step $t$	\\

                    $\mathcal{G}_{o_i}^t$                          &  Instantiation of the class template graph for  $o_i$ with class $C_j$ at time $t$\\
                    $\mathcal{G}^t_{\mathrm{inter}}$               & Dynamic object interaction graph at time step $t$\\
                    
			\bottomrule
		\end{tabular}
	\end{threeparttable}
\end{table}

\begin{figure}[t]
    \begin{subfigure}
     \centering
     \includegraphics[width=\textwidth]{fig_1.pdf}
\end{subfigure}
\begin{subfigure}
    \centering
\includegraphics[width=0.7\textwidth]{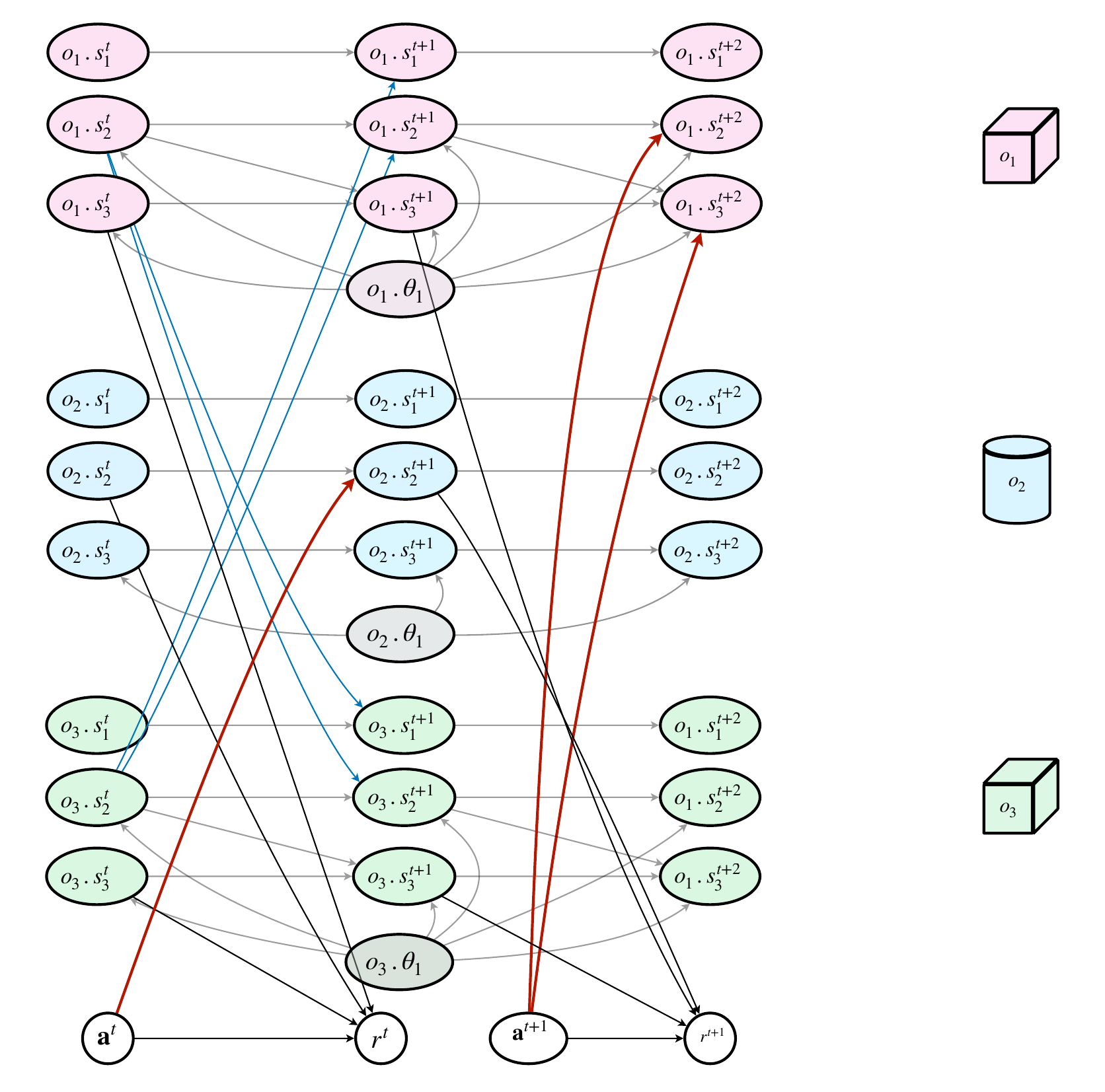}
    \end{subfigure}
        \caption{The graphical representation of DAFT-MDP. Fig A1.A represents the class template graphs for boxes and switches. Fig A1.B represents the interaction pattern graphs between two boxes. Fig A1.C represents the dynamic interaction graph, that at each timestep predicts which objects interact with each other and with the agent. The graph on the bottom shows an example of the instantiation of the complete ground graphical model for all of the objects in the environment from Fig A1.C, i.e. a pink box, a blue switch and a green box at time steps $t$, $t+1$ and $t+2$. The red lines describe the interactions of the agent with the objects (which follow the dotted lines in the class template graphs for each object class in Fig A1.A). The blue lines represent the interactions between objects, which follow the interaction patterns described in Fig A1.B.      \label{app-fig:factored_graph}}

\end{figure}

In this section, we first provide a summary of the notation in Table~\ref{app-table:notation}. Then we show an example of the environment described in the main paper, and how the learned graphs are connected in a single ground graphical model, as described in Figure~\ref{app-fig:factored_graph}.

In our example, we consider two classes 
$\mathcal{C} = \{C_1, C_2\}$, which represent the classes of boxes and switches, respectively. We represent these two classes with cubes and cylinders in Figure~\ref{app-fig:factored_graph}.

For the class of boxes $C_1$, represented in Figure~\ref{app-fig:factored_graph}A as a pink cube, we consider the attributes $\{C_1.s_1, C_1.s_2, C_1.s_3\}$ to represent the color, velocity and position of an abstract box in two consecutive timesteps $t$ and $t+1$. All of the relationship between the attributes are described in the class template graph $\mathcal{G}_{C_1}$. The edge $C_1.s_1^t \to C_1.s_1^{t+1}$ represents the fact that the color at timestep $t+1$ is only influenced by the color at timestep $t$ (in this case being constant). The edge $C_1.s_2^t \to C_1.s_2^{t+1}$ represents the fact that the velocity at timestep $t+1$ is influenced by the velocity at the previous timestep $t$. The edge $C_1.s_2^t \to C_1.s_3^{t+1}$ means that velocity can influence position in the next timestep, which is also influenced by the previous position $C_1.s_3^{t+1}$.
In this case the agent can optionally act (the dashed lines from $\mathbf{a}^t$) and influence the velocity $C_1.s_2^{t+1}$ and position $C_1.s_3^{t+1}$. Finally the latent constant parameters for the abstract box $C_1.\theta$, in this case representing friction, influence the velocity $C_1.s_2^{t+1}$ and position $C_1.s_3^{t+1}$ at each timestep $t+1$ and modulate the effect of the action on the position and velocity. The reward $r^t$ only depends on the position $C_1.s_3^{t}$. 

For the class of switches $C_2$, represented in Figure~\ref{app-fig:factored_graph}A as a blue cylinder,  we consider the attributes $\{C_2.s_1, C_2.s_2, C_2.s_3\}$ to represent the color, angle and position of an abstract switch in two consecutive timesteps $t$ and $t+1$. All of the relationship between the attributes are described in the class template graph $\mathcal{G}_{C_2}$. Similarly to boxes, the color $C_2.s_1$ does not influence anything and is only influenced by its previous value.
In this example, we consider that none of the attributes influences each other, but that the reward $r^t$ is only a function of the angle $C_2.s_2^t$. Additionally, the latent constant parameters $C_2.\theta$ only influence the position $C_2.s_3^t$, as can potentially the action $\mathbf{a}^t$ (represented by the dashed red lines).

In Figure~\ref{app-fig:factored_graph}B we show the interaction pattern graph for boxes $\mathcal{G}_{C_1,C_1}$ that represents the way that attributes of two objects of the class box $C_1$ interact with each other. In the figure, we use pink to represent the first object and green to represent the second object. Specifically, in this case, the interaction between two boxes means that the velocity of an object influences the velocity of another object. Similarly, the velocity of an object influences the color of the other object. Additionally, the latent constant parameters of an object influence the position of the other object.

In Figure~\ref{app-fig:factored_graph}C we consider a specific environment, specified by the objects $\mathcal{O}=\{ o_1, o_2, o_3\}$, where $ o_1$ and $ o_3$ are boxes, while $ o_2$ is a switch. We show an unrolled version of the dynamic interaction graph $\mathcal{G}_{\mathrm{inter}}$ for three consecutive timesteps for these objects. At timestep $t$ there is an interaction between the two boxes $ o_1$ and $ o_3$ (represented by blue lines), and the action $\mathbf{a}^t$ binds to the switch $ o_2$, as selected by the action selector $\boldsymbol{\alpha}^t$. The interactions between the two objects are instantiated following the interaction pattern graphs in Figure~\ref{app-fig:factored_graph}B. The action binding instead activates the red dashed lines in the class template graph in Figure~\ref{app-fig:factored_graph}A.
In the next timestep $t+1$ there are no interactions between objects, but the action is now bound to object $ o_1$.

In the bottom graph in Figure~\ref{app-fig:factored_graph}, we show how the three types of graphs are combined in this environment for these three consecutive timesteps. In particular, it can be seen that for each object, its attributes follow the same pattern as described by the class template graphs $\mathcal{G}_{C_1}$ and $\mathcal{G}_{C_2}$ (in grey). The interactions between the two boxes in timestep $t$ get instantiated following the interaction pattern graphs $\mathcal{G}_{C_1,C_1}$ (in light blue). The action binding specifies which object is affected by the action $\mathbf{a}^t$ at each timestep, but the way the action affects the attributes of an object is based on its class template graph. This graph represents the complete graphical model representation for these three timesteps, that describes how the transition and reward functions factorize in these three timesteps.

As can be seen from the example, to learn the factorization of the transition and reward functions in a new environment, we can reuse the class template graphs and the interaction pattern graphs across different domains with the same type of objects. We only need to relearn the dynamic interaction graph, which is also the only graph that is dynamic (i.e. the edges do not repeat across every couple of consecutive timesteps).

\section{Architecture and implementation details}\label{app:algorithm}

\begin{figure}[t]
     \centering
     \includegraphics[width=\textwidth]{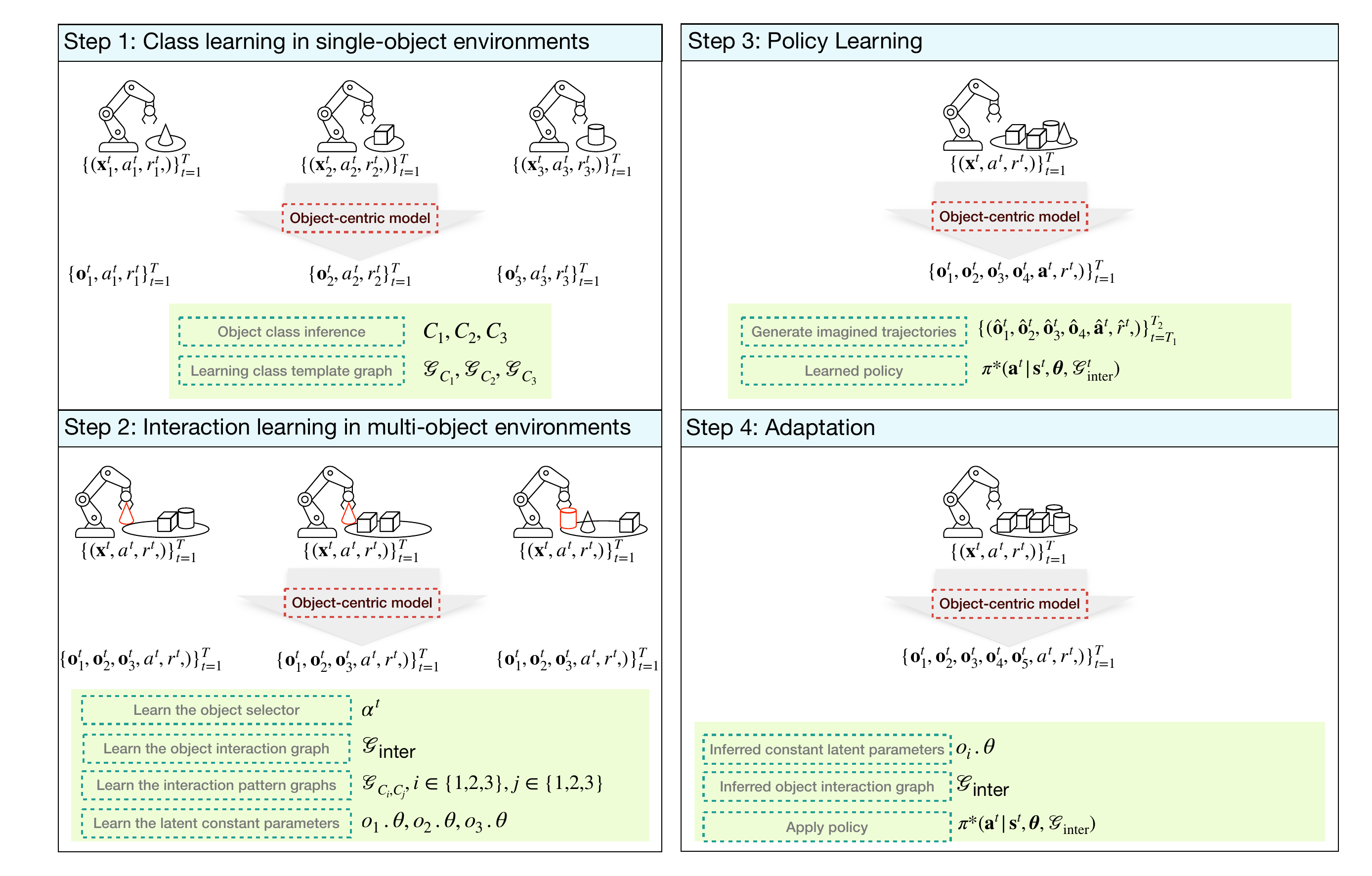}
        \caption{The learning pipelines of the DAFT-RL framework. }
    \label{app-fig:architecture}
\end{figure}

In this section, we describe the architecture of DAFT-RL in more detail.
Figure~\ref{app-fig:architecture} summarizes the pipeline of learning DAFT-RL framework, which uses the notation summarized in Table~\ref{app-table:notation}. We first provide a high-level illustration of the framework, and then provide the details in the following subsections.
In the offline model estimation, the DAFT-MDP is learned through a two-step process, involving the estimation of both the class template graphs (Step 1), the interaction pattern graphs and dynamic interaction graphs (Step 2). These graphs capture the relationships and dependencies among objects from the attribute level in the environment.  Once the DAFT-MDP is learned, policy learning is performed using trajectory imagination and planning methods (Step 3). During the adaptation phase, the dynamic graphs are inferred for the new domain (Step 4). This inference step allows for the transfer of the previously trained policy to the new domain without extensive retraining. By adapting the dynamic graphs to the new domain, the policy can be directly deployed and applied to the new environment. In the following, we define the graph structures and losses for each of these steps.

\subsection{Binary matrices in DAFT-RL}

We first introduce the definitions of all structures that are used in the loss functions in DAFT-RL:

\paragraph{Binary matrices in class template graphs.} As described in the previous sections, we learn the class template graphs $\mathcal{G}_{C_j}$ for each object class $C_j$. This graph is assumed to be constant in time. In practice, these graphs are learned as binary adjacency matrices. To help express the loss functions in a factorized way, we define the following binary matrices.

For $l=\{1, \dots, n\}$ we define:
\begin{itemize}
    \item  $C_j.G_{s \rightarrow s_l} \in \{0,1\}^n$ as the subset of the adjacency matrix of $\mathcal{G}_{C_j}$ that describes the edges incoming into $C_j.s_l^{t+1}$ from each $C_j.s_i^t$ for $i \in \{1, \dots, n\}$. In other words this binary vector is 1 only for the components $C_j.s_i^t$ for $i \in \{1, \dots, n\}$ that have an edge to $C_j.s_l^{t+1}$ in $\mathcal{G}_{C_j}$.
    
    \item  $C_j.G_{a \rightarrow s_l} \in \{0,1\}$ as a binary value that represents the existence of an edge between $\mathbf{a}^t$ and $C_j.s_l^{t+1}$, denoting how the action will affect the dynamics at the attribute $s_l$ at next time step.

    \item $C_j.G_{\theta \rightarrow s_l} \in \{0,1\}$ as a binary value that represents the existence of an edge between $C_j.\mathbf{\theta}^t$ and $C_j.{s}_l^{t+1}$ 
    
\end{itemize}
We additionally define:
\begin{itemize}  
    \item $C_j.G_{s \rightarrow r}\in \{0,1\}^n$ is the subset of the adjacency matrix of $\mathcal{G}_{C_j}$ that describes the edges incoming into $r^{t+1}$ from each $C_j.s_i^t$ for $i \in \{1, \dots, n\}$

    \item $C_j.G_{a \rightarrow r}\in \{0,1\}$ is a binary value that represents the existence of an edge between  $\mathbf{a}^t$.
\end{itemize}
For each object class $C_j$ all of these five groups of binary matrices together represent the adjacency matrix of graph $\mathcal{G}_{C_j}$.

\paragraph{Binary matrices in interaction pattern graphs.} Interaction pattern graphs $G_{C_i, C_j}$ represent the interaction structures between attributes of the object with class $C_i$ and the object with class $C_j$, which are assumed to be constant in time. 
For each $l=\{1, \dots, n\}$ we  define $G_{C_i \to C_j.s^l} \in \{0,1\}^n$ to represent the binary vector that describes  how the attributes of an object with class $C_i$ influence the attribute $C_j.s_l^{t+1}$ of the object with class $C_j$. All of these vectors together represent the adjacency matrix of graph $G_{C_i, C_j}$.

\paragraph{Binary values in Dynamic interaction graphs} The dynamic interaction graphs $G^t_{\mathrm{inter}}$ represent the interactions between the objects $\mathcal{O}$ at timestep $t$, which include the action selector $\{\boldsymbol{\alpha}^1, \boldsymbol{\alpha}^2, \ldots, \boldsymbol{\alpha}^T\}$, where $\boldsymbol{\alpha}^i \in \{0,1\}^m$ represents which object is affected by $\mathbf{a}^t$ at all time steps $t= 1, \dots, T$. We use the action selection $\boldsymbol{\alpha}^i$ in our losses.
For each $o_i, o_j \in \mathcal{O}$ we define a binary value $G^t_{\mathrm{inter}(i,j)} \in \{0,1\}$ to represent whether object $o_i$ and $o_j$ will have the interaction at time step $t$. All of these values together represent the adjacency matrix of graph $G^t_{\mathrm{inter}}$.

\subsubsection{Detailed loss functions}
We provide the detailed loss functions for learning the DAFT-MDP below. In each step, we mark in blue which of the structures defined in the previous subsections are learn at that step. 
\paragraph{Step 1: Class learning in single-object environments} As described in Sec~\ref{sec:step1}, we learn the class template graphs $\{\mathcal{G}^{\mathrm{C_1}}, \mathcal{G}^{\mathrm{C_2}}, \ldots, \mathcal{G}^{\mathrm{C_k}}\}$ using the collected $\mathcal{D}^{\mathrm{single}}$. We give the detailed loss functions, including the prediction loss $\mathcal{L}^1_{\mathrm{pred}}$ and the sparsity regularization $\mathcal{L}^1_{sparse}$ for dynamics and reward models, for this step below:

\begin{equation}
\begin{split}
    \mathcal{L}^1_{\mathrm{pred}} = \sum_{t=1}^T \sum_{i=1}^m \sum_{l=1}^n ( 
    & \log p_{{\color{blue}\lambda_s}}(o_i.s_l^{t+1} \mid {\color{blue} C_j.G_{s \rightarrow s_l}} \odot \mathbf{o_i}^t, {\color{blue} C_j.G_{a \rightarrow s_l}}  \odot \mathbf{a}_i^t) \\
    + & \log p_{{\color{blue}\lambda_r}}(r_i^t \mid {\color{blue} C_j.G_{s \rightarrow r}} \odot \mathbf{o_i}^t, {\color{blue}C_j.G_{a \rightarrow r}} \odot \mathbf{a}_i^t ))\\
\end{split}
\end{equation}

\begin{equation}
    \mathcal{L}^1_{\text{sparse}} = \sum_{j=1}^{k} \left( \sum_{l=1}^{n}\left\|{\color{blue} C_j.G_{{s} \rightarrow {s_l}}}\right\|_1 + \left\|{ \color{blue} C_j.G_{{s} \rightarrow {r}}}\right\|_1 + \sum_{l=1}^{n} \left\| {\color{blue} C_j.G_{a \rightarrow s_l}} \right\|_1 + \left\| {\color{blue} C_j.G_{a \rightarrow r}} \right\|_1 \right)
\end{equation}

where we update the prediction models including $\lambda_s, \lambda_r$, as well as the binary vectors defined previously $\{\{C_j.G_{s \rightarrow s_l}\}_{l=1}^n, \{C_j.G_{a \rightarrow s_l}\}_{l=1}^n, C_j.G_{s \rightarrow r}, C_j.G_{a \rightarrow r}\}$ (marked in blue). The complete loss $\mathcal{L}^1$ for Step 1 is:
\begin{equation}
    \mathcal{L}^1 = \mathcal{L}^1_{\mathrm{pred}} + \gamma^1_{\mathrm{sparse}} \mathcal{L}^1_{\text{sparse}}
\end{equation}
where $ \gamma^1_{\mathrm{sparse}}$ is a hyper-parameter.


\paragraph{Step 2.1: Learning the action binding} In this step, we learn the dynamic action selector $\boldsymbol{\alpha}^t$ by using the soft attention networks. We have collected $\mathcal{D}^{\mathrm{multi}}$ which have multiple objects in the scene. We consider that the class template graphs learned in the previous phase are fixed and update the prediction loss with the action selector. The detailed loss function for this stage is:
\begin{equation}
\label{l_2_pred_1}
\mathcal{L}^2_{\text {pred}}=\sum_{t=1}^{T} \sum_{i=1}^{m} \sum_{l=1}^n
\log p_{{\color{blue}\lambda_s}}\left(o_i.s_l^{t+1}\mid { C_i.G_{s \rightarrow s_l}} \odot \mathbf{o_i}^t, {C_i.G_{a \rightarrow s_l}}  \odot {\color{blue} \alpha_i^t} \cdot f_v(\mathbf{a}^t)\right)
\end{equation}
where $f_v$ is the value network to be updated. We also need to learn the key and query networks $f_k$ and $f_q$ to update  $\alpha_i^t$ for each object $o_i$ (marked in blue). Additionally, we use the same dynamic network $\lambda_s$ and also update it in this stage.

\paragraph{Step 2.2: learning dynamic interaction graph} 
As described in Sec~\ref{sec:step2.2}, we use  dNRI~\cite{graber2020dynamic} to learn the dynamic interaction graph. Additionally, we also learn all the interaction pattern graphs, as well as learn how to infer latent parameters, and how they influence each attribute, at this step.
We consider the class template graphs and action binding are fixed from the previous step.
Detailed loss functions are: 

\begin{equation}
\begin{split}
       \mathcal{L}^3_{\text {pred}}=
       & \sum_{t=1}^{T} \sum_{i=1}^{m} \sum_{l=1}^{n} \log p_{{\color{blue}\lambda_s}}(o_i.s_l^{t+1}\mid  {C_i.G_{s \rightarrow s_l}} \odot \mathbf{o_i}^t, { C_i.G_{a \rightarrow s_l}}  \odot {\alpha_i^t} \cdot f_v(\mathbf{a}^t),  \\
    & \qquad \qquad \qquad \quad \  {\color{blue} C_i.G_{\theta \rightarrow s_l}} \odot {\color{blue}o_i.\theta}, \{{\color{blue} G^t_{\text{inter} (i,j)}}\cdot {\color{blue} G_{C_j \rightarrow C_i.s_l}}\odot \mathbf{o}^{t}_j\}_{j\in\{1, \ldots, m\}\backslash i} )
\end{split}
\label{l_3_pred}
\end{equation}

\begin{equation}
    \mathcal{L}_\text{KL}^3 = \sum_{t=2}^T \text{KL} \left(q_{\color{blue} \phi}\left(\mathbf{z}^t \mid \mathbf{s}^{1:T}\right) \| p_{\color{blue}\phi}\left(\boldsymbol{z}^t \mid \boldsymbol{s}^{1: t}, \boldsymbol{z}^{1: t-1}\right)\right)
\end{equation}

\begin{equation}
    \mathcal{L}_\text{sparse}^3 = \sum_{i=1}^{k}\sum_{j=1}^{k}\sum_{l=1}^{n} \left\|{\color{blue}G_{C_i \rightarrow C_j.s_l}}\right\|_1  + \sum_{i=1}^{k} \sum_{l=1}^{n} \left\|{\color{blue}C_i.G_{\theta \rightarrow s_l}}\right\|_1
\end{equation}

where $\lambda_s$ is the dynamics model we have employed in the previous steps. We learn the encoder/prior networks $\phi$ to generate the latent vectors $\mathbf{z}$, where we sample the graph $G_\mathrm{inter}$ from. Through the binary vectors we also learn the interaction pattern graphs $\{G_{C_i,  C_j}\}_{i, j \in \{1, \ldots, k\}}$, latent parameters $\{o_i.\theta\}_{i=1}^m$ and the binary vectors representing the edges from latent parameters to each attribute $\{C_i.G_{\theta \rightarrow s_l}\}_{i=1}^k$. We have complete the loss for step 2.2:
\begin{equation}
    \mathcal{L}^3 = \mathcal{L}^3_{\text {pred}} + \gamma^3_{\mathrm{KL}} \mathcal{L}_\text{KL}^3 + \gamma^3_{\mathrm{sparse}} \mathcal{L}_\text{sparse}^3
\end{equation}
where $\gamma^3_{\mathrm{KL}}$ and $\gamma^3_{\mathrm{sparse}}$ are the hyper-parameters. 

\subsubsection{Details for the other algorithms}
\paragraph{Object class learning} For the case with pixel observation as input, we choose to learn the object classes in a supervised manner, where the input includes the extracted feature vectors $\mathbf{o}_i$ for each object and we have the labels $y_i$ for each single objects. We apply a cross-entropy loss to learn the mapping. 
\paragraph{Interaction graph sampling}
During the learning of dynamic interaction graphs, we generate the edge distribution $\mathbf{z}^t$ at each time step $t$. We sample the edge matrix $\mathbf{M}^t$ of the graph $\mathcal{G}_{\mathrm{inter}}^t$. Specifically, $\mathbf{M}^t \sim \mathrm{Bern}(\mathbf{z}^t)$, where $\mathrm{Bern}$ is the multivariate Bernoulli distribution and all elements are mutually independent. We also employ the Gumbel-Softmax trick~\cite{jang2017categorical} to make the sampling process differentiable.

\section{Experimental Details}
In this section we summarize all the experimental details. We first discuss our baselines and how we modified them to evaluate them in a fair comparison with DAFT-RL, focusing in particular on adapting symbolic input approaches to pixel inputs and image-based approaches to symbolic inputs. We then describe the experimental setups, including each task description and the modifications we made to the benchmarks. 

\subsection{Baseline modifications}\label{app:exp_baseline}
\paragraph{Symbolic input approaches with pixel inputs.} We adapt various approaches that rely on symbolic inputs, such as self-attention~\cite{zhou2022policy}, deep sets~\cite{zhou2022policy}, GNN~\cite{li2020towards}, SRICS~\cite{zadaianchuk2022self} and LRN~\cite{mambelli2022compositional}, to handle scenarios where the inputs are raw pixels. To accomplish this, we leverage pre-trained object-centric models, specifically AIR~\cite{eslami2016attend}, to obtain the state representation of each object within the scene.
The pre-trained object-centric models, specifically AIR, are employed to extract object-specific information from the raw pixel inputs. These object factored states, which represent the individual objects, are then used as inputs for the approaches mentioned above.
By incorporating the object-centric models and leveraging their extracted object representations, we enable the symbolic approaches originally designed for other input formats to be applicable in scenarios where raw pixels serve as the input representation.
\paragraph{Image-based approaches with symbolic inputs.} To modify the image-based approaches to fit with the benchmarks with symbolic inputs, we adopt the following changes:
\begin{itemize}
\item SMORL~\cite{zadaianchuk2021selfsupervised}: we remove the SCALOR encoder and directly obtain $\mathbf{z}$ from the simulator.

\item STOVE~\cite{Kossen2020Structured}: similarly, we remove the SuPAIR model and directly observe object-factored states for dynamics and policy modeling. 

\item 
NCS~\cite{chang2023hierarchical}: we directly observe the type and state variables from the simulator without the slot attention models. 
\end{itemize}
Notably, we do not and cannot modify  COBRA~\cite{watters2019cobra} for the symbolic case, because COBRA is mostly built upon the scene encoder and decoder.

\subsection{Experimental setups} \label{app:exp_setup}
In this section, we give the task description for each of the three benchmarks and describe the detailed experimental settings for both model estimation and policy learning. 

\subsubsection{Task description}

\paragraph{OpenAI Fetch - Push and Switch.} In this benchmarks the tasks are to push $N$ boxes or 
flip $M$ switches. 
The agent can obtain both the object factored states and the global (agent) states. Specifically, the object states are the object's pose and velocity. For switches, the states are also the switch angle and the position. The action is the 3D continuous control of the end effector displacement and 1D control of the distance between gripper fingers. The goals are the target position for cubes (pushing task) and the target angle of the switches (switching task). In the experiment, we consider a dense reward, computed by the distance between the object's states and the desired states in the goals. 

\paragraph{Spriteworld.} There are four tasks in the Spriteworld benchmark, as listed below. We follow the task sets as well as the modifications in~\cite{yoon2023investigation}. The agent directly observes the 2D images as the observation states. 
\begin{itemize}
    \item \textbf{Object goal task}: the goal is to move the object to the target position without touching other distractor objects. The action includes four cardinal directions. A positive reward will be given if the goal is achieved.  
    \item \textbf{Object interaction task}: the goal is to push the target to a given location in the scene. The reward will be obtained if the agent pushes the target to the goal position and the action also includes four cardinal directions.
    \item \textbf{Object comparison task}: the agent needs to figure out which object is different from other objects and move this object to the given position. The reward will be given if the goal is achieved and the action also includes four cardinal directions.
    \item \textbf{Property comparison task}: similarly, the agent needs to find out the object with different color or shape from other objects. This task is generally more challenging as the agent needs to reason the property-level difference while the object comparison task only requires object-level reasoning. The reward and action are the same as the object comparison task. 
\end{itemize}

\paragraph{Stacking.} 

We follow the block-stacking experimental setups in~\cite{nakanointeraction}. In the block-stacking task, the agent can observe the images from the current scene and the goal. The action includes picking and placing the objects and coordinates~\cite{veerapaneni2020entity}. The goal is to stack $m$ objects with a given position. 

\subsubsection{Benchmark modifications}
\paragraph{Color, shape, friction and mass in Push \& Switch} 
To make the benchmark more realistic, we add two variables into the original states in the benchmark, the object color and shape, both are represented as one-hot vector. 
Additionally, we also use a different friction coefficient and mass for each object. We implement this by conducting a post-processing filter for each object in the simulator. Specifically, different friction coefficients
will result in slower velocity and action effects. Different masses will result in different action effects of the agent. 
During the training of the estimation model, we use the objects with masses and friction coefficients uniformly sampled from $\{4,6,8,10\}$ and $\{0.4, 0.6, 0.8, 1.0\}$ respectively. And during testing, the object masses and friction coefficients are sampled from $\{1,2,3,11,13\} \cup \{3,5,7,9\}$ and $\{0.1, 0.2, 1.1, 1.3\} \cup \{0.5, 0.7, 0.9\}$ respectively.

\paragraph{Unseen colors, shapes and numbers of objects in Spriteworld} To evaluate the generalization capabilities of our model, we take into account the number of unseen objects, shapes, and colors. We achieve this without directly modifying the benchmark, as it offers the flexibility to alter these properties.
During the training phase for model estimation, we use the colors green, red, and yellow, along with a varied number of objects ranging from $\{3, 5, 7\}$. We consider as shapes circles and boxes.
During the testing phase, we introduce new elements that were not seen during training. Specifically, we incorporate the color blue and triangle shapes. Additionally, the number of objects is extended to include $\{2, 4, 6, 8\}$. By considering these changes in colors, shapes, and object numbers, we aim to assess the model's ability to generalize to novel configurations and demonstrate its robustness.

\paragraph{Mass in Stacking} We adjust the mass of each block by modifying the action effects on the object, treating them as post-processing filters applied to the image frames. In particular, picking up a heavier object will require more time compared to selecting a lighter one due to gravity. Placing a heavier one will become faster than the lighter one. During training, we consider the masses ranging from $\{1,1.5,2,2.5\}$. During testing, we make the masses of the objects ranging from $\{0.5, 1.25, 2.25, 3\}$. 

\subsubsection{Offline model estimation}

In Step 1, we use the GRU model~\cite{chung2014empirical} as both the dynamics and reward prediction model to learn the class template graph. The hidden size of the GRU is set to $128$. Additionally, we incorporate two MLP layers before and after the GRU, each with a hidden size of $64$.
During training, we continue optimizing the model until the dynamics and reward models converge, employing the mean squared error (MSE) loss with the Adam optimizer.
In Step 2, we use MLP layers to parameterize the soft attention layers~\cite{biza2022binding}. Specifically, for the functions $f_k$, $f_q$, and $f_v$, the hidden sizes are all set to $128$, with only one hidden layer for each of them.
To learn the dynamic interaction graph, we follow the hyperparameters and model architectures outlined in~\cite{graber2020dynamic}. However, as opposed to them,  we use the previously learned GRU in Step 1 as the decoder. For various experiments, we adopt different hyperparameters for the GRU decoder. The training details for each experiment are provided below.

\paragraph{Push \& Switch}

In Step 1, we gather a total of $400$ trajectories for both boxes and switches ($200$ for each). Each trajectory consists of $100$ time steps, and the actions are generated using a random policy. The batch size for this stage is set to $80$, and the weighting parameter before the regularization term is 0.015. In this step, in each trajectory, there is only one single object and different trajectories have different objects with different types (e.g., box or cubes), different colors (e.g., blue, red, etc), and different latent parameters (e.g., friction coefficients). 
In Step 2, we collect $500$ trajectories, where each trajectory consists of $50$ time steps, for learning the action binding selector. The batch size for training the soft attention network is set to $100$.
Regarding the learning of dynamic interaction graphs, we use the same parameters and model architectures as described in~\cite{graber2020dynamic}. For the decoder, we reuse the decoder used in Step 1, which includes the interaction pattern graphs and the class template graphs. The balancing weights before the KL divergence and the sparsity regularization are set to $0.9$ and $0.2$, respectively.
During training, we set the learning rate to $2.5 \times 10^{-5}$, and we apply a decaying factor of 0.5 every $100$ training epochs.

\paragraph{Spriteworld} We adopt the same pre-trained procedures outlined in~\cite{yoon2023investigation} for the object-centric models. Specifically, for both SA (Slot Attention)~\cite{locatello2020object} and AIR (Attend, Infer, Repeat)~\cite{eslami2016attend}, we generate a dataset containing various objects with distinct colors and shapes. The dataset comprises $200$ trajectories, each consisting of $50$ time steps. Within each frame, there are typically 2-5 objects exhibiting a diverse range of colors and shapes.
For Step 1, we use the collected dataset of $300$ trajectories (with $50$ time steps each) to learn the class template graphs with different objects. Each trajectory has the transitions of one single object. 
The weighting parameter before the sparsity regularizer is set to $0.3$. The batch size is $64$ and the learning rate is $0.002$. 
In our Spriteworld scenario, dense reward signals are absent. Therefore, we solely focus on learning the graphs between states and actions for most steps, excluding those that achieve the goal. Additionally, there is no direct interaction among objects, which means we do not train Step 2 for inferring the interactions. This setup aligns with~\cite{yoon2023investigation} and is consistent across all the baseline methods we employ.

\paragraph{Stacking} We also pre-train the SA and AIR using the randomly collected dataset containing $200$ trajectories, each with $30$ time steps. The blocks are initialized with different colors in the dataset. To train Step 1, we collect the single-block dataset with $200$ trajectories, each with $30$ time steps. The balancing weight for sparsity regularization is $0.05$. For Step 2, we have the dataset with $300$ trajectories, each with $30$ time steps and multiple objects. We use the same set of hyper-parameters for learning the soft attention networks as Push \& Switch. For learning the dynamic interaction graph, we balanced weights before the KL divergence and the sparsity regularization is $0.2$ and $0.1$, respectively. 
We use the same learning rate to $2.5 \times 10^{-5}$, and we apply a decaying factor of 0.5 every $100$ training epochs.


\subsubsection{Policy learning and planning} For the Push \& Switch and Spriteworld experiments, we use PPO~\cite{schulman2017proximal} for all baselines with the trajectories generated by the learned dynamics and reward functions. The learning rate for PPO is $0.005$ and $0.003$ in these two benchmarks, respectively. The coefficient of the entropy term in PPO is $0.05$. The policy is parameterized with $3$ MLP layers of size $256$ and $128$ in both experiments. For Stacking, we follow all the MPC hyper-parameter settings in~\cite{nakanointeraction} for the planning. 


\subsection{Full results}\label{app:exp_res}
We provide the full results of the experiments in this section, including the quantitative results for all experiments, ablation studies, and the visualized graphs. 
\subsubsection{Quantitative results}
Table~\ref{tbl:ps-single} and~\ref{tab:sw-single} give the results of the single training task in Push \& Switch and Spriteworld benchmarks. Table~\ref{tab:sw-number} and~\ref{tab:sw-cs} provide the results of Spriteworld with 1) changing object numbers and 2) changing object colors and shapes simultaneously during the testing phase, respectively. 
\begin{table}[h]
\centering
\caption{Average success rate over 3 random seeds for Push \& Switch environments testing (single task training mode).  The numbers in bold highlight the top-performing method.}
\resizebox{\textwidth}{!}{%
\begin{tabular}{@{}llllllllll@{}}
\toprule
\multirow{2}{*}{\textbf{\begin{tabular}[c]{@{}c@{}}Experiment\\ Settings\end{tabular}}} & \multicolumn{9}{c}{\textbf{Method}}                               \\ 
\cmidrule(l){2-10}              & \ours (Symbolic) & Deep Set  & Self-attention & SRICS & GNN & STOVE & Self-attention & NCS & Relational RL  \\ \midrule
2-Push 
& {$0.959$ \textpm $0.031$} & $0.985$ \textpm $0.025$ & $0.943$ \textpm $0.024$ & $0.981$ \textpm $0.015$ & $0.923$ \textpm $0.047$ & $0.973$ \textpm $0.028$ & $0.968$ \textpm $0.036$ & \textbf{0.993 \textpm 0.013} & $0.916$ \textpm $0.054$\\
2-Switch 
& \textbf{0.982 \textpm 0.013} & $0.869$ \textpm $0.032$ & $0.954$ \textpm $0.015$ & $0.978$ \textpm $0.029$ & $0.931$ \textpm $0.028$ & $0.916$ \textpm $0.045$ & $0.943$ \textpm $0.016$ & $0.977$ \textpm $0.038$ & $0.945$ \textpm $0.042$\\
3-Push &
{$0.961$ \textpm $0.035$} & $0.753$ \textpm $0.031$ & $0.670$ \textpm $0.023$ & $0.931$ \textpm $0.027$ & $0.784$ \textpm $0.041$ & $0.954$ \textpm $0.033$ & \textbf{0.972 \textpm 0.019} & $0.893$ \textpm $0.024$ & $0.929$ \textpm $0.022$  \\
3-Switch &
\textbf{0.907 \textpm 0.066} & $0.879$ \textpm $0.077$ & $0.805$ \textpm $0.089$ & $0.732$ \textpm $0.095$ & $0.295$ \textpm $0.169$ & $0.640$ \textpm $0.105$ & $0.660$ \textpm $0.083$ & $0.831$ \textpm $0.112$ & $0.748$ \textpm $0.089$ \\
\bottomrule
\end{tabular}%
}
\label{tbl:ps-single}
\end{table}

\begin{table}[h]
\centering
\caption{Average success rate over 3 random seeds for Spriteworld environments training (single task training mode). The numbers in bold highlight the top-performing method.}
\label{tab:sw-single}
\resizebox{\textwidth}{!}{%
\begin{tabular}{@{}llllllllll@{}}
\toprule
\multirow{2}{*}{\textbf{\begin{tabular}[c]{@{}c@{}}Experiment\\ Settings\end{tabular}}} & \multicolumn{9}{c}{\textbf{Method}}                               \\ 
\cmidrule(l){2-10}              & \ours (SA+TFD) & \ours (AIR)  & SMORL & SRICS & GNN & STOVE & COBRA & NCS & LRN  \\ \midrule
Object Goal &
  $0.916$ \textpm $0.032$ & $0.920$ \textpm $0.037$ & $0.745$ \textpm $0.066$ & $0.784$ \textpm $0.058$ & $0.464$ \textpm $0.091$ & $0.643$ \textpm $0.049$ & $0.715$ \textpm $0.049$ & \textbf{0.925 \textpm 0.021} & $0.846$ \textpm $0.067$
   \\
Object Interaction &
 $0.909$ \textpm $0.068$ & $0.896$ \textpm $0.053$ & $0.758$ \textpm $0.063$ & $0.824$ \textpm $0.096$ & $0.396$ \textpm $0.146$ & $0.683$ \textpm $0.094$ & $0.746$ \textpm $0.073$ & \textbf{0.931 \textpm 0.061} & $0.812$ \textpm $0.115$ \\
Object Comparison &
  $0.917$ \textpm $0.065$ & $0.902$ \textpm $0.060$ & \textbf{0.923 \textpm 0.049} & $0.812$ \textpm $0.071$ & $0.476$ \textpm $0.128$ & $0.625$ \textpm $0.091$ & $0.738$ \textpm $0.057$ & $0.901$ \textpm $0.051$ & $0.693$ \textpm $0.097$ \\
Property Comparison & 
\textbf{0.930 \textpm 0.034} & $0.905$ \textpm $0.075$ & $0.918$ \textpm $0.088$ & $0.810$ \textpm $0.095$ & $0.369$ \textpm $0.174$ & $0.602$ \textpm $0.108$ & $0.732$ \textpm $0.083$ & $0.897$ \textpm $0.112$ & $0.644$ \textpm $0.091$ \\
\bottomrule
\end{tabular}%
}
\end{table}

\begin{table}[h]
\centering
\caption{Average success rate over 3 random seeds for Spriteworld environments testing (with unseen object numbers). The numbers in bold highlight the top-performing method.}
\label{tab:sw-number}
\resizebox{\textwidth}{!}{%
\begin{tabular}{@{}llllllllll@{}}
\toprule
\multirow{2}{*}{\textbf{\begin{tabular}[c]{@{}c@{}}Experiment\\ Settings\end{tabular}}} & \multicolumn{9}{c}{\textbf{Method}}                               \\ 
\cmidrule(l){2-10}              & \ours (SA+TFD) & \ours (AIR)  & SMORL & SRICS & GNN & STOVE & COBRA & NCS & LRN \\ \midrule
Object Goal &
\textbf{0.928 \textpm 0.038} & $0.899$ \textpm $0.043$ & $0.720$ \textpm $0.069$ & $0.766$ \textpm $0.064$ & $0.450$ \textpm $0.095$ & $0.672$ \textpm $0.052$ & $0.699$ \textpm $0.055$ & $0.883$ \textpm $0.028$ & $0.791$ \textpm $0.071$ \\
Object Interaction &
\textbf{0.900 \textpm 0.074} & $0.886$ \textpm $0.059$ & $0.735$ \textpm $0.067$ & $0.802$ \textpm $0.100$ & $0.380$ \textpm $0.142$ & $0.665$ \textpm $0.100$ & $0.728$ \textpm $0.077$ & $0.851$ \textpm $0.066$ & $0.789$ \textpm $0.119$ \\
Object Comparison &
\textbf{0.904 \textpm 0.071} & {$0.890$ \textpm $0.065$} & $0.876$ \textpm $0.054$ & $0.795$ \textpm $0.076$ & $0.459$ \textpm $0.129$ & $0.674$ \textpm $0.096$ & $0.724$ \textpm $0.061$ & $0.864$ \textpm $0.058$ & $0.800$ \textpm $0.102$ \\
Property Comparison &
\textbf{0.911 \textpm 0.070} & $0.875$ \textpm $0.081$ & $0.865$ \textpm $0.093$ & $0.782$ \textpm $0.099$ & $0.355$ \textpm $0.173$ & $0.680$ \textpm $0.113$ & $0.711$ \textpm $0.087$ & $0.875$ \textpm $0.116$ & $0.772$ \textpm $0.097$ \\
\bottomrule
\end{tabular}%
}
\end{table}

\begin{table}[h]
\centering
\caption{Average success rate over 3 random seeds for Spriteworld environments testing (with unseen object colors and shapes). The numbers in bold highlight the top-performing method.}
\label{tab:sw-cs}
\resizebox{\textwidth}{!}{%
\begin{tabular}{@{}llllllllll@{}}
\toprule
\multirow{2}{*}{\textbf{\begin{tabular}[c]{@{}c@{}}Experiment\\ Settings\end{tabular}}} & \multicolumn{9}{c}{\textbf{Method}}                               \\ 
\cmidrule(l){2-10}              & \ours (SA+TFD) & \ours (AIR)  & SMORL & SRICS & GNN & STOVE & COBRA & NCS & LRN  \\ \midrule
Object Goal &
\textbf{0.902 \textpm 0.036} & $0.897$ \textpm $0.041$ & $0.694$ \textpm $0.071$ & $0.738$ \textpm $0.062$ & $0.420$ \textpm $0.099$ & $0.654$ \textpm $0.050$ & $0.681$ \textpm $0.053$ & $0.835$ \textpm $0.024$ & $0.775$ \textpm $0.069$ \\
Object Interaction &
\textbf{0.895 \textpm 0.072} & $0.880$ \textpm $0.057$ & $0.710$ \textpm $0.065$ & $0.772$ \textpm $0.102$ & $0.350$ \textpm $0.140$ & $0.635$ \textpm $0.098$ & $0.698$ \textpm $0.075$ & $0.859$ \textpm $0.064$ & $0.769$ \textpm $0.117$ \\
Object Comparison &
$0.878$ \textpm $0.069$ & \textbf{0.895 \textpm 0.063} & $0.854$ \textpm $0.052$ & $0.765$ \textpm $0.074$ & $0.429$ \textpm $0.125$ & $0.631$ \textpm $0.092$ & $0.704$ \textpm $0.059$ & $0.824$ \textpm $0.056$ & $0.724$ \textpm $0.100$ \\
Property Comparison &
\textbf{0.912 \textpm 0.068} & {$0.867$ \textpm $0.079$} & $0.885$ \textpm $0.091$ & $0.742$ \textpm $0.097$ & $0.325$ \textpm $0.171$ & $0.660$ \textpm $0.109$ & $0.690$ \textpm $0.085$ & $0.809$ \textpm $0.114$ & $0.725$ \textpm $0.093$ \\
\bottomrule
\end{tabular}%
}
\end{table}

\newpage
\subsubsection{Full ablation studies}
\begin{figure}[h]
     \centering
     \includegraphics[width=\textwidth]{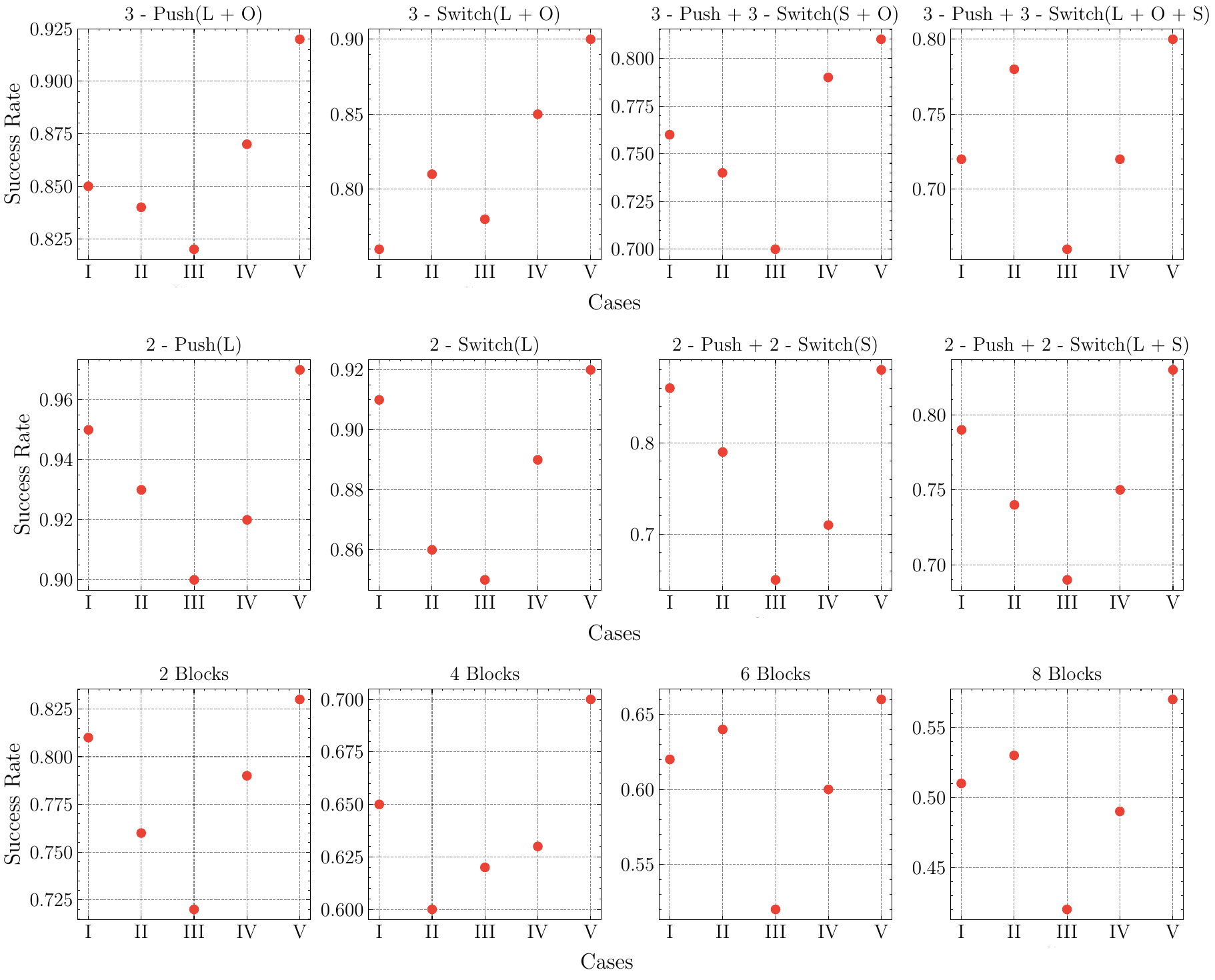}
        \caption{Ablation studies on Push \& Switch and stacking tasks. I. DAFT-RL w/o latent parameters; II. DAFT-RL w/o factored class template graph; III. DAFT-RL w/o dynamic interaction graph; IV. DAFT-RL w/o factored interaction pattern; V. Original DAFT.}
    \label{app-fig:abalation}
\end{figure}

Fig.~\ref{app-fig:abalation} gives the full ablation studies on Push \& Switch and Stacking benchmarks. We consider the following cases:
\begin{itemize}
    \item  DAFT-RL w/o latent parameters; 
    \item DAFT-RL w/o factored class template graph; 
    \item DAFT-RL w/o dynamic interaction graph; 
    \item DAFT-RL w/o factored interaction pattern.
\end{itemize}

Based on the obtained results, we observe that each learned component plays a crucial role in optimizing the policy. Notably, the dynamic interaction graph consistently demonstrates a significant contribution across the majority of experiments in Push \& Switch and stacking benchmarks. 

For Spriteworld, since there is no object interaction and latent parameters in this task, we only conduct the ablation studies on the class template graph. The results (Table~\ref{tbl:aba_sw}) suggest that the class template graph can benefit policy optimization as well. 
\begin{table}[h]
\scriptsize
\centering
\caption{Ablation studies on the Spriteworld benchmark.}
\label{tbl:aba_sw}
\begin{threeparttable}
\begin{tabular}{c|c|c}
\toprule
Task & DAFT (SA) & DAFT (SA) w/o class template graphs \\ \hline
Object goal & $0.897$ & $0.767$  \\ \hline 
Object interaction & $0.890$ & $0.739$  \\ \hline 
Object comparison & $0.893$ & $0.685$  \\ \hline 
Property comparison & $0.907$ & $0.832$  \\ 
\bottomrule
\end{tabular}
\end{threeparttable}
\end{table}

These findings highlight the importance of factored and dynamic graphs in capturing and modeling the attribute-level interactions between objects. They suggest that understanding and incorporating the dynamic relationships and dependencies among objects have a substantial impact on policy optimization.

\subsubsection{Visualization of the learned graphs}
\begin{figure}[h]
     \centering
     \includegraphics[width=\textwidth]{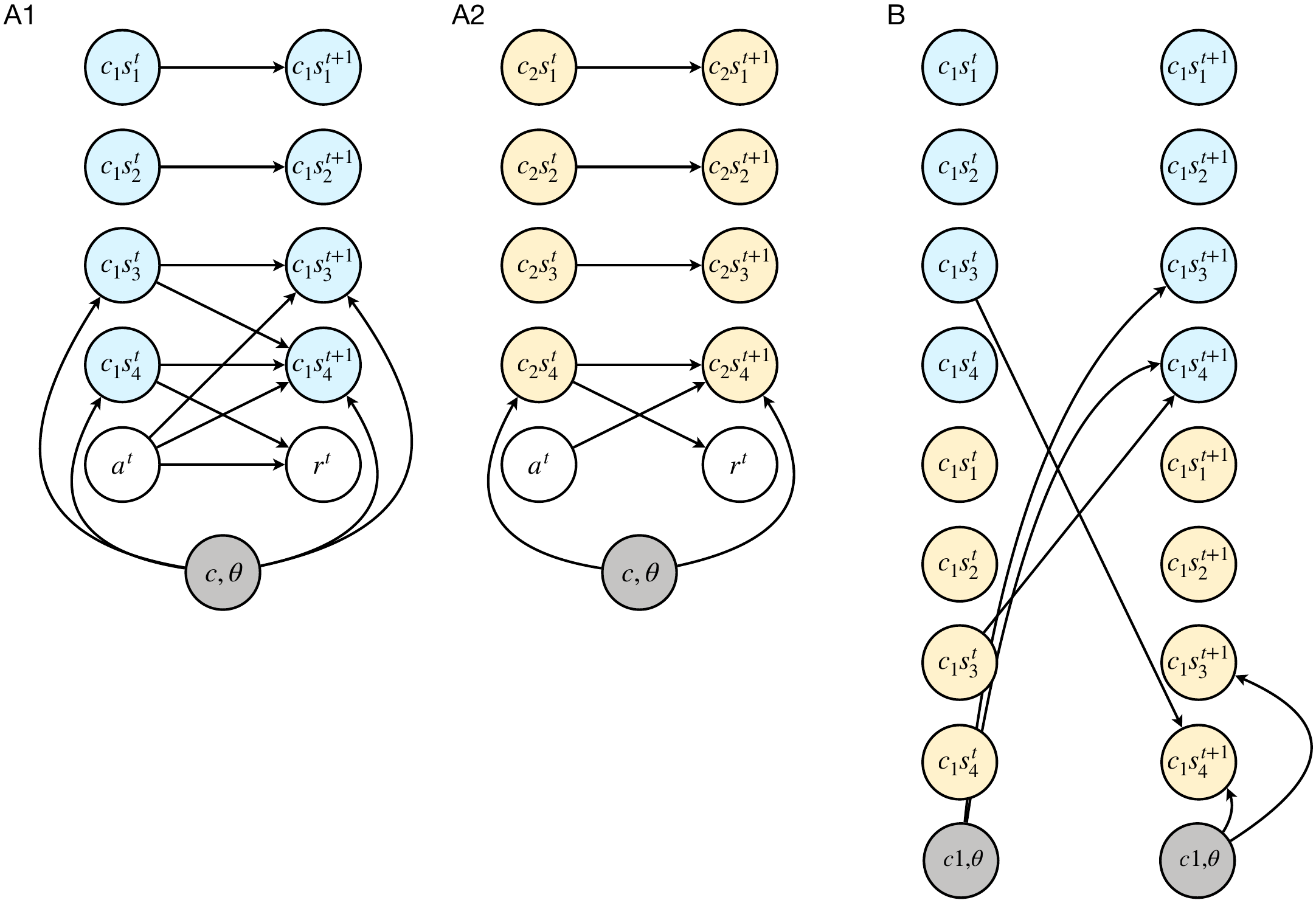}
        \caption{Learned class template graph and interaction pattern graph in push \& switch task. A1-A2: Class template graphs for box and switch. B: interaction pattern graph among two boxes.}
    \label{app-fig:graph_vis}
\end{figure}

In Figure \ref{app-fig:graph_vis}, which illustrates the learned graphs for boxes and switches in the Push \& Switch benchmarks, we denote the class template graphs for boxes and switches as A1 and A2. 
The variables $C_1.s_1$, $C_1.s_2$, $C_1.s_3$, and $C_1.s_4$ represent the color, shape, velocity, and position of the boxes, respectively. Similarly, $C_2.s_1$, $C_2.s_2$, $C_2.s_3$, and $C_2.s_4$ correspond to the color, shape, position, and angle of the switches. In both cases, $C_i.\theta$ and $C_j.\theta$ indicate the friction coefficients. 

As expected, in the learned class template graph, we observe that shape and color do not have a direct effect on the dynamics, and they remain unchanged throughout the dynamics. On the other hand, velocity directly influences the position of the boxes, while the position itself does not affect the velocity.
Regarding the switches, their positions do not directly impact the angles since they are fixed at specific positions on the panel. Also, as expected, actions applied to the switches have an effect solely on their angles.
The reward is based on the actions, positions, and angles for both boxes and switches.
Additionally, the latent parameters influence the velocity and position of the boxes, while they specifically affect the angle of the switches.

As for the interaction graph between two boxes, shown in Figure \ref{app-fig:graph_vis}B, the velocity of one box can impact the position of the other box, while the other variables do not have a direct influence on each other. This learned interaction graph successfully captures the underlying physical processes of the system, enabling an attribute-level understanding of the object interactions.
Overall, these learned graphs effectively recover the underlying physical processes of the system.

\end{document}